\documentclass{article}
\usepackage{xr}
\usepackage[utf8]{inputenc}
\usepackage{graphicx}
\usepackage{amsmath}
\usepackage{wrapfig}
\usepackage{makecell}
\usepackage[makeroom]{cancel}
\usepackage{algorithm2e}
\usepackage{algorithmic}
\usepackage{xcolor}
\usepackage{tikz} 
\usetikzlibrary{positioning,arrows.meta,quotes,bayesnet,matrix,calc}
\usepackage{xcolor}
\usepackage{caption}
\usepackage{subcaption}
\usepackage{capt-of}
\usepackage{bbm}
\usepackage{amssymb}
\usepackage{stackrel}
\usepackage{pifont}
\usepackage[numbers]{natbib}

\newcommand{\bracket}[1]{\mathopen{}\left[ {#1}_{{}_{}}\,\negthickspace\right]\mathclose{}}
\newcommand{\set}[1]{\mathopen{}\left\{ {#1}_{{}_{}}\,\negthickspace\right\}\mathclose{}}

\newcommand{\E}{\mathbbm{E}}

\usepackage{arxiv}

\usepackage[utf8]{inputenc} % allow utf-8 input
\usepackage[T1]{fontenc}    % use 8-bit T1 fonts
\usepackage{hyperref}       % hyperlinks
\usepackage{url}            % simple URL typesetting
\usepackage{booktabs}       % professional-quality tables
\usepackage{amsfonts}       % blackboard math symbols
\usepackage{nicefrac}       % compact symbols for 1/2, etc.
\usepackage{microtype}      % microtypography
\usepackage{lipsum}		% Can be removed after putting your text content
\usepackage{graphicx}
\usepackage{natbib}
\usepackage{doi}
\author{
  Elouan Argouarc'h, Fran\c{c}ois Desbouvries, \\
  SAMOVAR \\
  Télécom SudParis, Institut Polytechnique de Paris \\
   91120 Palaiseau France\\
  \texttt{\{elouan.argouarch, francois.desbouvries\}@telecom-sudparis.eu} \\
  \And
  Eric Barat, Eiji Kawasaki, \\
  CEA-List\\
  Université Paris-Saclay \\
   91120 Palaiseau France\\
  \texttt{\{eric.barat, eiji.kawasaki\}@cea.fr} \\
}
\begin{document}
\title{Generative vs. discriminative modeling under the lens of uncertainty quantification}
\maketitle
\begin{abstract}
Learning a parametric model from a given dataset indeed enables to capture intrinsic dependencies between random variables via a parametric conditional probability distribution and in turn predict the value of a label variable given observed variables. In this paper, we undertake a comparative analysis of generative and discriminative approaches which differ in their construction and the structure of the underlying inference problem. Our objective is to compare the ability of both approaches to leverage information from various sources in an epistemic uncertainty aware inference via the posterior predictive distribution. We assess the role of a prior distribution, explicit in the generative case and implicit in the discriminative case, leading to a discussion about discriminative models suffering from imbalanced dataset. 
We next examine the double role played by the observed variables in the generative case, and discuss the compatibility of both approaches with semi-supervised learning. We also provide with practical insights and we examine how the modeling choice impacts the sampling from the posterior predictive distribution. With regard to this, we propose a general sampling scheme enabling supervised learning for both approaches, as well as semi-supervised learning when compatible with the considered modeling approach. Throughout this paper, we illustrate our arguments and conclusions using the example of affine regression, and validate our comparative analysis through classification simulations using neural network based models.
\end{abstract}

\section{Introduction}\label{introduction}

The statistical learning tasks of classification 
and regression \cite{vapnik2013nature}\cite{hastie2009elements} are paramount in many scientific fields and have gained increasing interest in the big-data and machine learning era. Elaborate methods and tools \cite{baydin2018automatic}\cite{bottou2018optimization} enable to leverage flexible parametric models in order to capture intrinsic dependencies between related variables, and in turn predict the value of a variable of interest given observed values of the others.

Many statistical learning methods, both historical \cite{aitken1936least}\cite{berkson1944application}\cite{morgan1963problems} and recent \cite{murphy2012machine}\cite{li2021survey}\cite{murphy2012machine}, can be understood as building an approximate of the conditional probability distribution of a variable of interest (which we refer to as label in this paper) given the value of an observed variable. This is usually achieved by considering a parameterized model and adjusting the parameters by minimizing a loss function computed on a dataset comprised of recorded couples of observations and labels. 

In this context, two main approaches are often opposed and compared \cite{ng2001discriminative}\cite{ulusoy2005generative}\cite[\S 9.4]{murphy2012machine} \cite{lueckmann2021benchmarking}\cite{fetaya2019understanding}\cite{zheng2023revisiting}. The discriminative modeling consists in parameterizing a distribution over the label given the observations as in \cite{lecun1998gradient}{\cite{krizhevsky2009learning}}\cite{greenberg2019automatic} for example; while the generative modeling techniques parameterize a distribution over the observed variable given the label, see \cite{revow1996using}\cite{papamakarios2017masked}\cite{papamakarios2019sequential} \cite{mackowiak2021generative} for some applications.

In many learning tasks, the observations are high-dimensional random variables (rv) when compared to the usually low-dimensional labels. Therefore, discriminative models are much more convenient to work with compared to generative models as the later involve modeling a high-dimensional conditional distribution. However, recent developments in generative probabilistic modeling \cite{goodfellow2020generative}\cite{papamakarios2021normalizing}\cite{sohl2015deep} which now enable to capture intrinsic distributions have paved the way towards a renewed appeal of the generative modeling approach \cite{odena2016semi}\cite{izmailov2020semi}\cite{zimmermann2021score}.

When predicting a label of interest using a parametric model (be it generative or discriminative) from an observation, committing to a unique value can lead to high imprecision as there are two sources of uncertainty that one needs to account for \cite{der2009aleatory}\cite{kendall2017uncertainties}\cite{hullermeier2021aleatoric}. Uncertainty aware modeling aims at computing or estimating confidence/credible intervals associated to a prediction, but this task of uncertainty quantification remains challenging \cite{alaa2020discriminative}. The \textit{aleatoric} uncertainty is the source of uncertainty which results from the stochastic nature of the unknown (or at least with untractable probability density function (pdf)) Data Generating Process (DGP) which, as we assume, generates the observations from the labels. 
However, using an approximate parametric model also induces an additional uncertainty about the predicted labels, which is referred to as \textit{epistemic} \cite{huang2021quantifying}.

Bayesian modeling methods for uncertainty quantification are perhaps amongst the most promising approaches \cite{ghahramani2015probabilistic}: the model parameters are treated as random, and are marginalized-out to obtain the predicted law of label given the observation as well as the dataset.
This so-called {\sl posterior predictive distribution} (ppd)
will from now on be the distribution of interest, 
and our aim throughout this paper is to compute (or at least sample from) this distribution.

Such techniques include building conjugate models for which, by construction,
the model parameter posterior distribution is easy to sample from,
see e.g. 
\cite{raftery1997bayesian}
\cite{hannah2011dirichlet}. In the case of more elaborate (typically neural network (NN) based) models, prior conjugacy can be leveraged to some extent in methods such as Bayesian last layer \cite{fiedler2023improved},
but the ppd can no longer be sampled from with a straightforward procedure. 
Fortunately, 
Bayesian methods still enable to sample from this distribution \cite{neal2012bayesian}, be it via MCMC methods \cite{chen2016bridging}\cite{wenzel2020good}\cite{izmailov2021bayesian} or variational inference \cite{graves2011practical}\cite{gal2016dropout}\cite{rudner2022tractable}. 
Finally, approximate methods such as posterior Bootstrap \cite{newton1994approximate}\cite{lyddon2018nonparametric}\cite{fong2019scalable}\cite{newton2021weighted} do indeed provide with theoretically grounded approaches for Bayesian model averaging, or bagging, uncertainty quantification. 

Our aim in this paper is to compare the generative and discriminative approaches under the scope of Bayesian uncertainty quantification.

The rest of our paper is organized as follows.
In section \ref{supervised} we first provide with a precise description of both generative and discriminative constructions.
Next in section \ref{epistemic uncertainty}, by analysing the ppd, we explain different behaviours of the two modeling approaches in an epistemic uncertainty aware inference. More specifically, in section \ref{explicit_implicit_prior}, we focus our attention on the role of a specific distribution which can be understood as a prior distribution associated to the ppd, and analyze the ability of each approach to infer using prior information. By doing so, we give clues as to why discriminative models can suffer from imbalanced datasets, while the generative ones do not, 
and confirm this analysis via both illustrative and quantitative simulations. In order to sample from the ppd, especially in the generative case, we provide in section \ref{gibbs_supervised} with a general sampling algorithm which is based on a Gibbs scheme and which can easily be applied to both approaches. 
Finally in section \ref{semi-supervised}, we specifically discuss the dependency of model parameters to observations, and conclude on the compatibility of each approach with the task of semi-supervised learning which aims at inferring the model parameters from both labeled and unlabeled datasets. We propose to leverage the corresponding Gibbs sampling scheme to perform Bayesian semi-supervised learning in the generative case. We finally perform simulations in the context of image classification in order to illustrate the arguments of this paper in different learning scenarios. 

\section{Supervised learning: Context and objective}\label{supervised}

Let $(X,Y)$ be a couple of rv related via a DGP $\mathcal{P}_{Y|X}$ which describes the probability distribution of $Y$ given $X$. The task of prediction consists in retrieving information about an unknown $x_0$ which (is assumed to have) generated an observed $y_0$ via the DGP: $y_0 \sim \mathcal{P}_{Y|X}(Y|X=x_0).$
In this paper, we use the specific nomenclature of a classification problem and we denote $Y$ as \emph{observation} (from the DGP) and, $X$ as the corresponding \emph{label} (though it is not necessarily categorical). The Bayes formula tells us that, once the value $y_0$ is observed, the information about $x_0$ is encapsulated in the \emph{posterior} distribution $\mathcal{P}_{X|Y}$ with probability density function\footnote{throughout this paper, the pdf should be understood w.r.t. the appropriate measure, depending on the nature (continuous, discrete or mixed continuous-discrete) of the underlying variables.} (pdf) given by: 
\begin{equation}
    p_{X|Y}(x_0|y_0) = \frac{p_{Y|X}(y_0|x_0)\pi_{X_0}(x_0)}{p_{Y_0}(y_0)}, \text{where }
    p_{Y_0}(y_0) = \int p_{Y|X}(y_0|x)\pi_{X_0}(x) \mathrm{d}x,  \label{posterior}
\end{equation}
where $p_{Y|X}({\bf .}|{\bf .})$ is the conditional pdf
associated with the DGP and $\pi_{X_0}({\bf .})$ is the pdf associated with the distribution which describes our \emph{prior} knowledge about $x_0$ (in this paper we denote
prior distributions with letter $\Pi$
and their pdf with $\pi$). This posterior distribution can be used to obtain pointwise Bayes predictors by minimizing the expectation of a well-designed loss function $l$ \cite{lehmann2006theory}: $x_0^* = \arg\min_{x_0}\E_{X|Y}\bracket{l(X,x_0)};$ but in essence, this posterior distribution describes our inability to commit to a singular value of $x_0$. This source of uncertainty is induced by the random nature of the DGP and is referred to as \emph{aleatoric}.

If both of these pdf can be evaluated, 
then \eqref{posterior} can be computed at least up to the constant denominator $p_{Y_0}(y_0) = \int p_{Y|X}(y_0|x)\pi_{X_0}(x)\mathrm{d}x$. In many situations however, the pdf associated with the DGP is intractable and consequently \eqref{posterior} cannot be evaluated, not even up to a constant. This situation occurs either when (i) we only dispose of a dataset $\mathcal{D}$ (defined in the next paragraph) generated from  and the DGP (and so its pdf) is otherwise simply unknown; or (ii) when the DGP is only available via its stochastic simulation procedure which enables to obtain (and augment \cite{durkan2018sequential}\cite{lueckmann2019likelihood}) a dataset $\mathcal{D}$ but its pdf is implicit \cite{cranmer2020frontier}; historical approaches in this setting include the Approximate Bayesian Computation (ABC) methods \cite{csillery2010approximate}. In this paper we consider the first setting and suppose that we dispose of $\mathcal{D}$ generated from the DGP but that we no longer have access to the random sampling mechanism of the DGP making ABC unfeasible. A possible approach to cope with this shortcoming is to resort to an approximation of the intractable posterior using a conditional probability distribution $\mathcal{P}_{\theta}$ where parameter $\theta$ is inferred using observed couples $\mathcal{D} \stackrel{\Delta}{=} \{(x_i, y_i)|x_i \sim \mathcal{P}^\mathcal{D}_{X}, y_i\sim \mathcal{P}_{Y|X}(Y|X = x_i)\}_{i=1}^{|\mathcal{D}|}$. We denote $\mathcal{P}_{X}^{\mathcal{D}}$ the probability distribution which effectively generated the values in dataset $\mathcal{D}$ but we stress here that this probability distribution is not necessarily the same as the prior distribution $\Pi_{X_0}$ (this particular point and its consequences are discussed in details in section \ref{explicit_implicit_prior}). This general formulation includes the usual tasks of statistical parametric learning: we talk about regression (resp. classification) when $X$ is continuous (resp. categorical). 

Since a unique parameter estimate of $\theta$ (such as Maximum Likelihood Estimates (MLE) or Maximum A Posteriori (MAP) estimates) can be stained with high imprecision if $\mathcal{D}$ is not representative enough of the DGP, we rather consider $\theta$ to be a hidden rv, assume a prior knowledge described by distribution $\Pi_\Theta$, and approximate \eqref{posterior} with the ppd \cite{gelman1995bayesian}:
\begin{equation}\label{posterior_predictive}
    p(x_0 | y_0, \mathcal{D})
    = \int p(x_0,\theta|y_0,\mathcal{D})\mathrm{d}{\theta}.
\end{equation}
This pdf indeed accounts for the \emph{epistemic} uncertainty which is the uncertainty about the unknown parameter $\theta$ (induced by the finite number of recorded samples in $\mathcal{D}$) propagated to $x_0$ predicted by this model. The ppd  \eqref{posterior_predictive} is computed by integrating out $\theta$ in the joint pdf: 
\begin{equation}\label{joint}
    p(x_0, \theta|y_0,\mathcal{D})  = p(x_0|y_0, \theta)p(\theta|y_0,\mathcal{D}),
\end{equation}
and so $p(x_0|y_0, \mathcal{D}) = \mathbbm{E}_{\theta}[p(x_0|y_0,\theta)|y_0, \mathcal{D}]$ which explains the denomination: it is the average of (posterior) \emph{predictions} $p(x_0|y_0,\theta)$ over probable models under the \emph{posterior} $p(\theta|y_0,\mathcal{D})$. Ultimately, computing the posterior pdf \eqref{posterior_predictive}, or sampling from the distribution if exact computation of the expectation by integration is unfeasible, would allow to identify the probable (aleatoric) values of $x_0$ that might have generated $y_0$ via the DGP, while accounting for the modeling (epistemic) uncertainty. 

\subsubsection*{\sl Illustrating running example}
\begin{figure}[!ht]
    \centering
    \includegraphics[width=.7\textwidth]{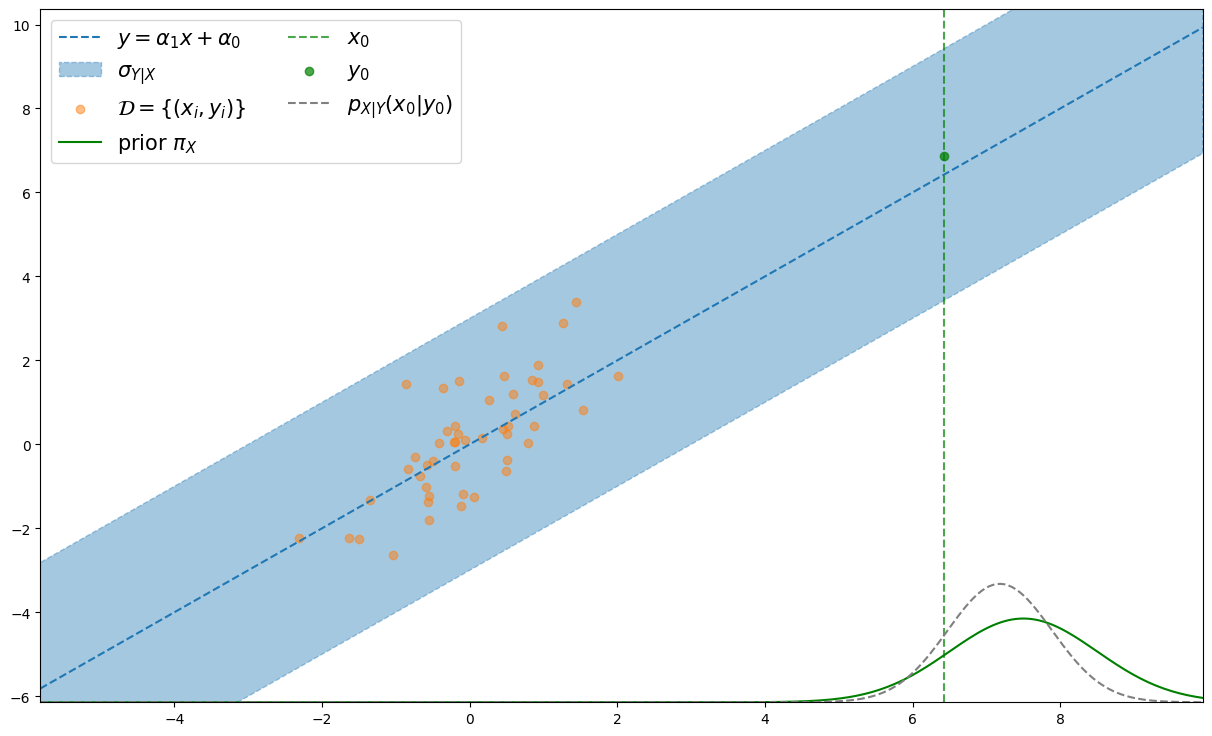}
    \caption{Supervised learning setting}
\label{fig:supervised_affine_setting}
\end{figure}

Throughout this paper, we propose to illustrate arguments and conclusions on the continued example of affine regression. Though it serves as an illustrative example, this specific application to affine modeling is, in itself, relevant since affine relationship between variables of interest are most frequent in many science fields and are in many cases, the first considered dependency hypothesis. 
Let $(X,Y)\in \mathbb{R}\times\mathbb{R}$ 
be two real-valued rv (assumed to be) related via an unknown DGP. 
For illustration purposes, 
we consider a toy setting where the DGP is of the form: 
\begin{equation}\label{true_DGP_example}
    Y = \alpha_1X + \alpha_0 + \sigma_{Y|X}\epsilon \iff p_{Y|X}(y|x) = \mathcal{N}(y;\alpha_1 x + \alpha_0, \sigma^2_{Y|X}).
\end{equation}
We dispose of recorded data $\mathcal{D}$ produced by the DGP, and where the distribution of the $x$ values is $\mathcal{P}_X^{\mathcal{D}}$. The prior knowledge 
on $x_0$ is given by the distribution $\pi_{X_0}(x_0) = \mathcal{N}(x_0; \mu_{X_0},\sigma_{X_0}^2)$. We chose the prior and the DGP to be conjugated such that the posterior distribution reads:
\begin{equation}\label{true_posterior_example}
    p_{X|Y}(x_0|y_0) = \mathcal{N}(x_0; (\frac{\alpha_1^2}{ \sigma^2_{Y|X}} + \frac{1}{\sigma_{X_0}^2})^{-1}(\frac{\alpha_1(y_0 - \alpha_0)}{ \sigma^2_{Y|X}} + \frac{\mu_{X_0}}{\sigma_{X_0}^2}), (\frac{\alpha_1^2}{ \sigma^2_{Y|X}} + \frac{1}{\sigma_{X_0}^2})^{-1});
\end{equation}
and so this posterior distribution can be used to assess the quality of the inference when comparing the ppd $p(x_0|y_0,\mathcal{D})$ to it; but we otherwise suppose the DGP unavailable. An example of this setting is illustrated in figure \ref{fig:supervised_affine_setting}.

\subsection{Generative versus Discriminative modeling}
\begin{figure}[!ht]
    \centering
    \includegraphics[width=.9\linewidth]{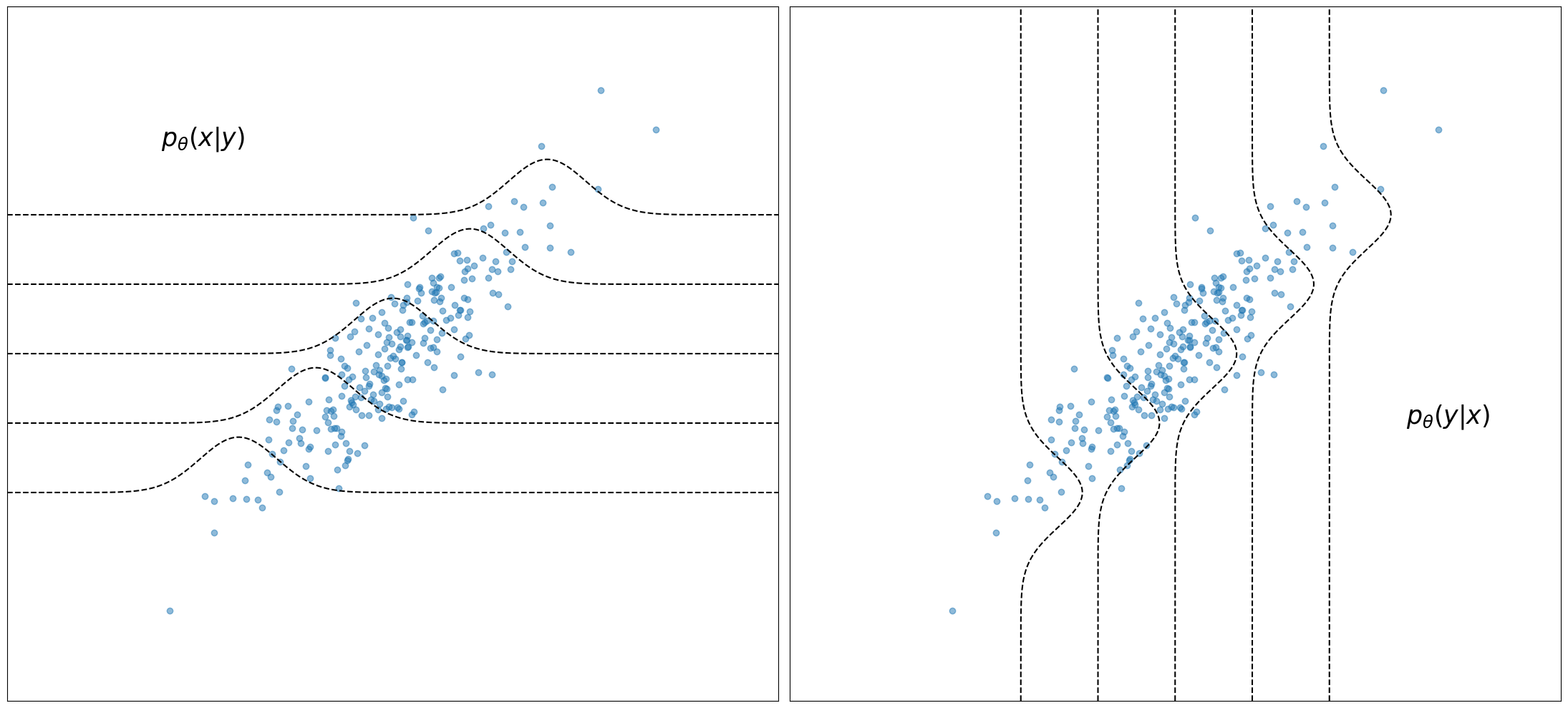}
    \caption{Illustration of the difference between generative and discriminative modeling approaches (right)}
    \label{fig:illustration}
\end{figure}

In the previous section we explained the general principle of modeling the posterior pdf \eqref{posterior}, and we emphasized on the role of the ppd \eqref{posterior_predictive}, which accounts for the epistemic modeling uncertainty. However, we have not explained precisely yet how the modeling is carried out. In fact, using a parametric conditional probability distribution $\mathcal{P}_\theta$, we can either model the unknown DGP with $\mathcal{P}_\theta(Y|X)$ and deduce the corresponding posterior via the Bayes formula, or model the posterior directly with $\mathcal{P}_\theta(X|Y)$. 
In the literature, the first approach is classically referred to as \emph{Generative} (since it models the data \emph{generating} process), while the second one is called  \emph{Discriminative} (since it makes sense in particular in the  classification setting, where the model directly computes the label probabilities which enable to \emph{discriminate} samples via their respective classes). The first approach is called generative modeling but in the (deep) Machine Learning literature, generative modeling \cite{bond2021deep} can also refer to the task building a parametric probability distribution which is \textit{generative} in the sense that it can be sampled from easily and is designed to resemble a probability which produced recorded data. In this paper and unless stated otherwise, generative modeling refers to the approach which consists in building an approximate of the posterior distribution of interest \eqref{posterior} via modeling the unknown likelihood.
These two approaches differ in their philosophy: the first one models only what is unknown, i.e. the generative process, while the second one directly models the function of interest, i.e. the posterior pdf. Figure \ref{fig:illustration} provides with an illustration of the difference between the two approaches. 

These approaches yield two different equations: 
\begin{align}
    &p(x_0|y_0,\theta)
    \stackrel{G}{=}
    \frac{p_\theta(y_0|x_0)\pi_{X_0}(x_0)}{\int p_\theta(y_0|x)\pi_{X_0}(x)\mathrm{d}x}\label{generative_posterior} \\
    &p(x_0|y_0,\theta)
    \stackrel{D}{=}
    p_\theta(x_0|y_0),\label{discriminative_posterior}
\end{align}
in which superscripts $G$ and $D$  respectively stand for generative and discriminative; this notation will be used throughout the rest of this paper. In both equations, 
$p_\theta$, be it $p_\theta(y_0|x_0)$ in the generative case or $p_\theta(x_0|y_0)$ in the discriminative case, is the conditional pdf associated with $\mathcal{P}_\theta$, and is what is effectively computed with model associated with parameter $\theta$.  

\begin{figure}[!ht]
    \begin{subfigure}[b]{0.45\textwidth}
        \centering
        \begin{tikzpicture}[transform shape, node distance=1cm, roundnode/.style={circle, draw=black, very thick, minimum size=5mm},squarednode/.style={rectangle, draw=black, very thick, minimum size=5mm}][h]
            \node[obs](y_0){$y_0$};%
            \node[roundnode](x_0)[above= of y_0]{$x_0$};%
            \node[obs](y_1)[xshift = 2cm]{$y_1$};%
            \node[obs](y_N)[xshift = 4cm]{$y_{|\mathcal{D}|}$};%
            \node[obs](x_1)[above= of y_1]{$x_1$};%
            \node[obs](x_N)[above= of y_N]{$x_{|\mathcal{D}|}$};%
            \node[roundnode][below=of y_1](theta){$\theta$}; %
            \node at ($(y_1)!.5!(y_N)$) {\ldots};
            \node at ($(x_1)!.5!(x_N)$) {\ldots};
            \node[obs](y_0){$y_0$};%
            \edge {x_0,theta}{y_0}
            \edge {x_1,theta}{y_1} 
            \edge {x_N,theta}{y_N} 
        \end{tikzpicture}
    \end{subfigure}
    \hfill
    \begin{subfigure}[b]{0.45\textwidth}
        \centering
        \begin{tikzpicture}[transform shape, node distance=1cm, roundnode/.style={circle, draw=black, very thick, minimum size=5mm},squarednode/.style={rectangle, draw=black, very thick, minimum size=5mm}][h]
            \node[obs](y_0){$y_0$};%
            \node[roundnode](x_0)[above= of y_0]{$x_0$};%
            \node[obs](y_1)[xshift = 2cm]{$y_1$};%
            \node[obs](y_N)[xshift = 4cm]{$y_{|\mathcal{D}|}$};%
            \node[obs](x_1)[above= of y_1]{$x_1$};%
            \node[obs](x_N)[above= of y_N]{$x_{|\mathcal{D}|}$};%
            \node[roundnode][above=of x_1](theta){$\theta$}; %
            \node at ($(y_1)!.5!(y_N)$) {\ldots};
            \node at ($(x_1)!.5!(x_N)$) {\ldots};
            \node[obs](y_0){$y_0$};%
            \edge {y_0,theta}{x_0}
            \edge {y_1,theta}{x_1} 
            \edge {y_N,theta}{x_N} 
        \end{tikzpicture}
    \end{subfigure}
    \caption{Graphical models compared: Generative (left) versus Discriminative (right). The grey (resp. white) nodes are observed (resp. latent) variables.}\label{fig:graphical}
\end{figure}
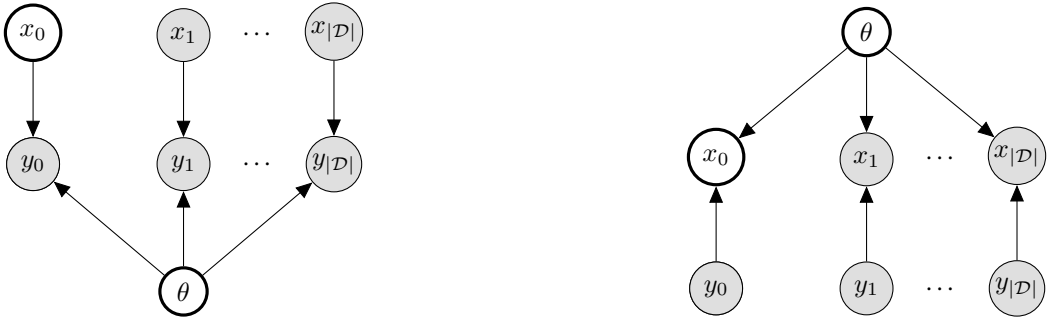

Figure \ref{fig:graphical} displays a graphical representation of both models and explains how the modeling choice affects the dependence between all the rv. We build upon these two figures by writing the full joint pdf of all rv. 
This comparison of graphical models allow us to deduce the joint distribution of all the rv of interest: 
\begin{eqnarray}
    % p(x_0,y_0, \mathcal{D}) = \pi_{X_0}(x_0)p_{Y|X}(y_0|x_0)\prod_{(x_i,y_i)\in\mathcal{D}} p^{\mathcal{D}}_{X}(x_i) p_{Y|X}(y_i|x_i)\\
    p(x_0,y_0,\theta,\mathcal{D}) \stackrel{G}{=} \pi_\Theta(\theta)\pi_{X_0}(x_0) p_\theta(y_0|x_0)\prod_{(x_i,y_i)\in\mathcal{D}} p^{\mathcal{D}}_{X}(x_i) p_\theta(y_i|x_i),\label{generative full joint} \\
    p(x_0,y_0,\theta,\mathcal{D}) \stackrel{D}{=} \pi_\Theta(\theta)p_{Y_0}(y_0)p_{\theta}(x_0|y_0)\prod_{(x_i,y_i)\in\mathcal{D}} p^{\mathcal{D}}_{Y}(y_i) p_{\theta}(x_i|y_i)\label{discriminative full joint}.
\end{eqnarray} 
The difference between generative and discriminative modeling has already been established in the literature. In the context of classification, \cite{ng2001discriminative} compares the Naive Bayes Classifier (which is a generative model) and logistic regression (which is discriminative) in term of (asymptotic) classification error and conclude in favor of the generative approach when working with a small amount of training data. 

% In this paper, we compare both modeling choices through the lens of uncertainty quantification. We first examine the aleatoric uncertainty by discussing the compatibility of both approaches with the underlying principles of Bayes formula \eqref{posterior}. We then explain how the epistemic uncertainty is accounted for in both approaches by sampling from the ppd $p(x_0|y_0,\mathcal{D})$. 

\subsubsection*{\sl Illustrative running example}
We now illustrate the difference between both modeling approaches by performing homoskedastic affine regression with unknown variance Gaussian Noise where model parameters are the coefficients of the affine transform as well as the variance of the unknown noise so $\theta = \{\beta_1, \beta_0, \sigma^2\}$. Both the DGP \eqref{true_DGP_example} and the posterior \eqref{true_posterior_example} belong to the considered parametric family of conditional probability, so both modeling approaches correspond to a well-specified inference problem. We first describe the generative approach $y_0 \stackrel{G}{=} \beta_1 x_0 + \beta_0 + \sigma\epsilon, \epsilon\sim \mathcal{N}(0,1) \iff p_\theta(y_0|x_0) \stackrel{G}{=} \mathcal{N}(y_0;\beta_1 x_0 + \beta_0, \sigma^2)$.
Once again, thanks to the conjugacy of the prior $\pi_{X_0}(x_0)$ and the previous $p_\theta(y_0|x_0)$, the posterior pdf $p(x_0|y_0,\theta)$ admits a closed-form expression:
\begin{equation}\label{x0_posterior_affine_generative}
    p(x_0|y_0,\theta) \stackrel{G}{=} \mathcal{N}(x_0; (\frac{\beta_1^2}{\sigma^2} + \frac{1}{\sigma_{X_0}^2})^{-1}(\frac{\beta_1(y_0 - \beta_0)}{\sigma^2} + \frac{\mu_{X_0}}{\sigma_{X_0}^2}), (\frac{\beta_1^2}{\sigma^2} + \frac{1}{\sigma_{X_0}^2})^{-1}).
\end{equation}
We now describe the discriminative modeling approach with the same parameterized $\mathcal{P}_\theta$: 
\begin{equation}\label{affine_discriminative_model}
    x_0 \stackrel{D}{=} \beta_1 y_0 + \beta_0 + \sigma\epsilon, \epsilon \sim \mathcal{N}(0,1) \iff p_\theta(x_0|y_0) \stackrel{D}{=} \mathcal{N}(x_0;\beta_1 y_0 + \beta_0, \sigma^2).
\end{equation}

\subsection{Handling multiple observations}\label{handling_mulitple_observations}
The bayesian philosophy behind equation \eqref{posterior} is to assume (i) prior knowledge on $x_0$ (before observation) in the form of a prior $\Pi_{X_0}$  and (ii) an observation model. This observation model is assumed to produce $y_0$ from $x_0$ but it might not necessarily be true in practice. In this case, we say that the observation model is mispecified w.r.t. the observation. In our case, we considered the observation model to be the DGP $\mathcal{P}_{Y|X}$, so \eqref{posterior} is a well specified Bayesian setting. Then, after observing (one or) several observations: $y_{0,1},...,y_{0, N_0} \stackrel{iid}{\sim} \mathcal{P}_{Y|X}(Y|X=x_0)$.
We deduce the posterior pdf:
\begin{align}\label{multiple_observations}
p_{X|Y_{0,1},...,Y_{0,N_0}}(x_0|y_{0,1},...,y_{0, N_0}) &= \frac{\pi_{X_0}(x_0)\prod_{n=1}^{N_0}p_{Y|X}(y_{0,n}|x_0)}{\int \pi_{X_0}(x_0)\prod_{n=1}^{N_0}p_{Y|X}(y_{n}|x_0)\mathrm{d}y_1...\mathrm{d}y_{N_0}}.
\end{align}
We often describe this equation as a form of Bayesian updating: we update the prior knowledge with the observations. In section \ref{explicit_implicit_prior}, we will discuss the role of the prior $\Pi_{X_0}$ with regard to both modeling approaches; but in this section, we first specifically examine whether or not each approach enables to easily handle multiple observations in the inference of $x_0$. 

Equations \eqref{generative_posterior} and \eqref{discriminative_posterior} explain how we can predict the value of $x_0$ from a unique observed value of $y_0$ using model $\theta$ for respectively the generative and the discriminative approach. In this case, both approaches enable computation of the posterior $p(x_0|y_0,\theta)$ as both equations are tractable (at least up to a constant). By contrast, when we observe not only a single observation but rather a collection of observations from the DGP which originate from the same unknown value of interest, as in \eqref{multiple_observations}, then the generative approach allows to handle this situation with a tractable equivalent of \eqref{generative_posterior}, while the discriminative one does not.

Indeed, under a generative modeling, we can easily rewrite equation \eqref{generative_posterior} as: 
\begin{equation}\label{generative_posterior_multiple}
    p(x_0|y_{0,1},...,y_{0, N_0}, \theta) \stackrel{G}{=} \frac{\pi_{X_0}(x_0)\prod_{n=1}^{N_0}p_\theta(y_{0,n}|x_0)}{p(y_{0,1},...,y_{0, N_0}|\theta)};
\end{equation}
and this formula can be computed up to its constant denominator (w.r.t. $x_0$). On the other hand, with a discriminative modeling, equation \eqref{discriminative_posterior} becomes: 
\begin{equation}\label{discriminative_posterior_multiple}
    p(x_0|y_{0,1},..., y_{0,N_0}, \theta) \stackrel{D}{=} \frac{\prod_{n=1}^{N_0}p_{Y_n}(y_{0,n})p_\theta(x_0|y_{0,n})}{p(x_0|\theta)^{N_0-1}p_{Y_1,...,Y_{N_0}}(y_{0,1},...,y_{0,N_0})}. 
\end{equation}
However, 
factor $p(x_0|\theta)\stackrel{D}{=} \int p_\theta(x_0|y)p_{Y_0}(y)\mathrm{d}y$ is always intractable since $p_{Y_0}(y)$ given by \eqref{posterior} is defined implicitely by the unknown DGP. Therefore, 
\eqref{discriminative_posterior_multiple} cannot be evaluated, not even up to a constant, when $N_0>1$. Finally, only the generative approach allows to conveniently deal with multiple observations. In order to carry on with the comparison of both approaches, we only consider the case of a unique observation $y_0$, but, concerning the generative modeling, the equations of the rest of this paper still hold with multiple observations. 

\section{Supervised Epistemic Uncertainty via the ppd}\label{epistemic uncertainty}
We now discuss how the epistemic uncertainty is accounted for in each approach, be it generative or discriminative. To that end we analyze how the modeling choice impacts the ppd and more precisely how it can be sampled from. We proceed in three steps:  first we analyze the model posterior distribution $p(\theta|y_0,\mathcal{D})$ (see \S \ref{with_without_observation}), we then deduce the joint distribution $p(x_0,\theta|y_0,\mathcal{D}) \stackrel{\eqref{joint}}{=} p(x_0|y_0,\theta)p(\theta|y_0,\mathcal{D})$ (see \S \ref{derivation_of_joint}) and we finally come to its (other) marginal of interest, i.e. the ppd \eqref{posterior_predictive} (see \S \ref{sampling_from_the_posterior_predictive}).

\subsection{model posterior: $p(\theta|y_0,\mathcal{D})$ or $p(\theta|\mathcal{D}) ? $}\label{with_without_observation}

In this section we look at the posterior distribution over model $\theta$ given the observation $y_0$ and the recorded dataset $\mathcal{D}$. Using Bayes rule, it can be written as: 
\begin{equation}
    p(\theta|y_0, \mathcal{D}) =\frac{p(y_0|\theta,\xcancel{\mathcal{D}})}{p(y_0|\mathcal{D})}p(\theta|\mathcal{D}),
    \text{ where } p(\theta|\mathcal{D}) \propto \pi_\Theta(\theta) \prod_{(x_i,y_i)\in\mathcal{D}} p(x_i,y_i|\theta)\label{model_posterior_no_y0}.
\end{equation}
It is important to note here that in the previous equation we can cancel out $\mathcal{D}$ since for any variables involved in figure \ref{fig:graphical}, we have $p(.|., \theta,\mathcal{D}) = p(.|.,\theta)$. By glancing at these two equations, we can already see that the probable values of $\theta$ under this posterior correspond to models for which the elements of $\mathcal{D}$, the couples $(x_i,y_i)$, are likely. In this section, we discuss the impact of $y_0$ on this distribution and conclude on whether or not this observation carries information for inference of $\theta$ depending on the modeling approach. To that hand, we start by leveraging equations \eqref{generative full joint} and \eqref{discriminative full joint} to deduce:
\begin{eqnarray}
& p(y_0|\mathcal{D}) \stackrel{G}{=}\int p(y_0|\theta)p(\theta|\mathcal{D})\mathrm{d}\theta 
\text{ where } p(y_0|\theta) \stackrel{G}{=} \int p_\theta(y_0|x)\pi_{X_0}(x)\mathrm{d}x,
\label{y0_theta_D_generative}\\
&p(y_0|\theta) \stackrel{D}{=} p(y_0|\mathcal{D}) \stackrel{D}{=} p_{Y_0}(y_0).\label{y0_theta_D_discriminative}
\end{eqnarray}

On the one hand, with a generative approach, $p(y_0|\theta)$ indeed depends on $\theta$,
so $y_0$ indeed carries information for inferring of $\theta$ since the two rv are not independent. We can moreover analyze how the information is carried by $y_0$ to a posteriori models. Probable generative models $\theta$ under posterior $p(\theta|y_0,\mathcal{D})$ produce, with high probability, the value $y_0$ for unknown values $x$ distributed under the prior $\Pi_{X_0}$. The posterior distribution of models $\theta$ therefore effectively depends on $y_0$. We finally conclude that the role of $y_0$ in the ppd inference is twofold: (i) in conjonction to the prior $\Pi_{X_0}$ it indeed carries information to probable (epistemic) models $\theta$ and (ii) it carries information to probable (aleatoric) $x_0$ values via posterior models $\theta$. 

On the other hand, under a discriminative approach, 
factors $p(y_0|\theta)$ and $p(y_0|\mathcal{D})$ reduce to $p_{Y_0}(y_0)$ (see \eqref{y0_theta_D_discriminative}) so $\theta$ and $y_0$ are independent rv and finally the posterior over $\theta$ reduces to $p(\theta|\mathcal{D})$. Let us analyze why observation $y_0$ does not carry any information to a posteriori models. The information carried by $y_0$ to a discriminative model $\theta$ is that it should produce, with high probability, unknown values $x$ for $y_0$. However, this is nothing but saying that $p_\theta(.|Y = y_0)$ is a probability distribution, which we already know by construction of a discriminative model using $\mathcal{P}_\theta$ a conditional probability distribution. So, by contrast with the generative approach, in the discriminative approach, the role of $y_0$ is solely aleatoric, i.e. to infer $x_0$ via probable discriminative models which do not depend on $y_0$.

\subsection{Joint pdf}\label{derivation_of_joint}

We now derive the joint pdf $p(x_0,\theta|y_0,\mathcal{D})$ given by equation \eqref{joint} for both generative and discriminative modeling approaches. In the generative case \eqref{generative_posterior}, 
as explained before, the first factor in the joint pdf is $p(x_0|y_0, \theta) \stackrel{G}{=} p_\theta(y_0|x_0)\pi_{X_0}(x_0)/p(y_0|\theta)$. In general, this expression can only be computed up to a normalizing constant since $p(y_0|\theta) = \int p_\theta(y_0|x)\pi_{X_0}(x)\mathrm{d}x$ might be intractable. However, this denominator is a constant w.r.t. $x_0$ but it indeed depends on $\theta$ so it must not be treated as a constant in the joint pdf; so, with regard to the joint pdf, the first factor cannot be computed. Moreover,  the second factor is $p(\theta|y_0,\mathcal{D})\propto p(y_0|\theta)p(\theta|\mathcal{D})$, and as we have explained before in the previous section \ref{with_without_observation} indeed depends on $y_0$. For the same reason, $p(y_0|\theta)$ is intractablethe second factor in the joint pdf cannot be computed, not even up to a constant, either. Conveniently, both factors are intractable because of the same factor $p(y_0|\theta)$ which appears in the denominator of the first and in the numerator of the second. So, even though none of the two factors can be computed individually, the intractable terms cancel out by multiplication and the joint pdf can be computed up to a constant (w.r.t. both $x_0$ and $\theta$) as: 
\begin{align}
    p(x_0, \theta|y_0, \mathcal{D}) &\stackrel{G}{=}
    \frac{\pi_{X_0}(x_0)p_\theta(y_0|x_0)}{p(y_0|\theta)}
    \frac{p(\theta|\mathcal{D})p(y_0|\theta)}{p(y_0|\mathcal{D})} 
    \stackrel{G}{\propto} \pi_{X_0}(x_0)p_\theta(y_0|x_0)p(\theta|\mathcal{D}) \label{generative_joint}\\
    \text{where } p(\theta|\mathcal{D})&\stackrel{G}{\propto}\pi_{\Theta}(\theta)\prod_{(x_i,y_i)\in \mathcal{D}} p_{\theta}(y_i|x_i)\label{generative_model_posterior}.
\end{align}
In the discriminative setting, the first factor in the joint pdf \eqref{joint} reads $p(x_0|y_0,\theta) = p_\theta(x_0|y_0)$ (see again equation \eqref{discriminative_posterior}). This quantity is directly computed in a normalized way by model $\mathcal{P}_\theta$. Moreover, as we pointed out in the previous section \ref{with_without_observation}, the second factor reduces to $p(\theta|\mathcal{D})$ which can be computed up to a constant. So, unlike in the generative case, both factors can be computed and the joint pdf therefore reads: 
\begin{align}
    p(x_0, \theta|y_0, \mathcal{D}) &\stackrel{D}{=}  p_\theta(x_0|y_0)p(\theta|\mathcal{D})\label{discriminative_joint}\\ 
    \text{where } p(\theta|\mathcal{D}) &\stackrel{D}{\propto} \pi_\Theta(\theta) \prod_{(x_i,y_i)\in\mathcal{D}}p_{\theta}(x_i|y_i)\label{discriminative_model_posterior}.
\end{align}
So in both modeling cases, the joint pdf can be computed (at least up to a constant).

\subsection{The ppd}\label{sampling_from_the_posterior_predictive}
Recall that the ppd \eqref{posterior_predictive} is obtained by marginalizing out the rv $\theta$ in the joint distribution \eqref{joint}. 
The consequence is twofold: first its pdf is obtained by integrating the joint pdf w.r.t. variable $\theta$; and second, sampling from the joint distribution provides $x_0$ samples which are distributed under the ppd. In this section, we discuss the first point. 

By integrating 
\eqref{generative_joint} and \eqref{discriminative_joint} w.r.t. $\theta$, we obtain expressions for the ppd \eqref{posterior_predictive} in both cases: 
\begin{align}\label{generative_posterior_predictive}
    &p(x_0|y_0, \mathcal{D})\stackrel{G}{\propto} \pi_{X_0}(x_0)\int p_{\theta}(y_0|x_0)p(\theta|\mathcal{D})\mathrm{d}\theta\\
    \label{discriminative_posterior_predictive}
    &p(x_0|y_0, \mathcal{D}) \stackrel{D}{=} \int p_\theta(x_0|y_0) p(\theta|\mathcal{D})\mathrm{d}\theta
\end{align}
In practice, these two equations can only be used
when exact computation of the integral is feasible.
Nonetheless, they remain relevant 
as we can analyze them both to grasp a difference between the two approaches which provides with another interpretation of the ppd. The first formula, in the generative case, corresponds to Bayesian inference using the prior and the marginal likelihood $p(y_0|x_0,\mathcal{D}) = \int p_{\theta}(y_0|x_0)p(\theta|\mathcal{D})\mathrm{d}\theta$; while the second formula, in the discriminative case, corresponds to an averaging of posterior predictions. 
So, in both cases, the ppd is an averaging of  the quantity $p_\theta$ (which is what is effectively computed with model $\theta$) w.r.t. $p(\theta|\mathcal{D})$. This is to be contrasted with the original definition of the ppd, defined as the average of predictions w.r.t. $p(\theta|y_0,\mathcal{D})$. These two definitions only coincide in the discriminative case since the model computes directly the prediction (see \eqref{discriminative_posterior}) and the posterior distribution $p(\theta|y_0,\mathcal{D})$ reduces to $p(\theta|\mathcal{D})$ (see section \ref{with_without_observation}). As a consequence, this means that $\int p(x_0|y_0,\theta)p(\theta|\mathcal{D})\mathrm{d}\theta$ is an exact construction of the ppd only in the discriminative case. 

\subsection{Explicit or Implicit prior}\label{explicit_implicit_prior}
A main difference between the two approaches lies in the role of the $x_0$ marginal in the joint pdf $p(x_0,y_0|\mathcal{D})$. This distribution is of particular interest as it can be considered as a prior $p(x_0|\mathcal{D})$ in the ppd $p(x_0|y_0,\mathcal{D})$ which is the object of interest in both modeling approach:
\begin{equation}\label{implicit_prior}
    p(x_0|y_0,\mathcal{D}) \propto p(x_0|\mathcal{D})p(y_0|x_0,\mathcal{D}). 
\end{equation}
We can again leverage both equations \eqref{generative full joint} and \eqref{discriminative full joint} to deduce: 
\begin{align}
        &p(x_0|\mathcal{D}) \stackrel{G}{=} p(x_0|\theta) \stackrel{G}{=} \pi_{X_0}(x_0);\label{x0_theta_D_generative} \\ 
        &p(x_0|\mathcal{D}) \stackrel{D}{=} \int p(x_0|\theta)p(\theta|\mathcal{D}) \mathrm{d}\theta\text{ where }p(x_0|\theta) \stackrel{D}{=} \int p_\theta(x_0|y)p_{Y_0}(y)\mathrm{d}y\label{x0_theta_D_discriminative}.
\end{align}
These two equations allow to understand that the $x_0$ marginal does not play the same role in the generative and discriminative cases. While in the former setting this immutable marginal distribution describes prior knowledge and does not depend on $\mathcal{D}$ (see equation \eqref{x0_theta_D_generative}); in the latter setting, this marginal distribution is the result of an intricate interaction between the dataset $\mathcal{D}$, the prior distribution $\Pi_{X_0}$ and the DGP $\mathcal{P}_{Y|X}$. On the one hand, in the generative approach, it corresponds to the prior $\Pi_{X_0}$ which can be specified according to the problem at hand and, in itself, may provide significant information about the value of interest $x_0$. In practice, this prior distribution can also play a role of regularization and may as well be understood as a safeguard since it can effectively constrain the prediction to a specific region of the space \cite{williams1995bayesian}\cite{steck2002dirichlet}\cite{piironen2017sparsity} but more importantly, the prior distribution is often the result of an elicitation effort \cite{robert2007bayesian} which consists in of (i) obtaining prior information and (ii) transcribing this knowledge into a probability distribution. 
On the other hand with a discriminative approach, this $x_0$ marginal has a very different role. The relevance of equation \eqref{x0_theta_D_discriminative} first lie in the fact that it highlights the systematic intractablity of pdf $p(x_0|\mathcal{D})$. Indeed, it can never be computed (even if exact integration was feasible) since it ultimately involves computing the pdf $p_{Y|X}$ in $p_{Y_0}(y) = \int p_{Y|X}(y|x)\pi_{X_0}(x)\mathrm{d}x$, which is unknown by assumption. This intractability does not pose any practical issue since the computation of the $x_0$ marginal $p(x_0|\mathcal{D})$ \eqref{x0_theta_D_discriminative} is not required for computing the joint pdf \eqref{discriminative_joint} (and consequently not for computing the pdf of (or sampling from) the ppd). But this also means that the discriminative construction does not allow to leverage any information encapsulated in, or any practical property induced by, a prior distribution during the inference. We now rewrite the expression of equation \eqref{x0_theta_D_discriminative} as:     
\begin{equation}
    p(x_0|\mathcal{D}) \stackrel{D}{=} \int\int\int p_\theta(x_0|y)p(\theta|\mathcal{D})p_{Y|X}(y|x)\pi_{X_0}(x)\mathrm{d}x\mathrm{d}y\mathrm{d}\theta
    \stackrel{D}{=} \int p(x_0|y,\mathcal{D})p_{Y_0}(y)\mathrm{d}y\label{sampling_from_x0_given_D}.
\end{equation}

So, this distribution has a pdf $p(x_0|\mathcal{D})$ which indeed depends on (i) $\mathcal{D}$ via the unknown $\theta$ and is indirectly related to (ii) the prior $\Pi_{X_0}$ and (iii) the DGP  $\mathcal{P}_{Y|X}$ via $\mathcal{P}_{Y_0}$. As a consequence, this distribution will attribute high probability mass to the $x$ values which have high probability under $p(x_0|y,\mathcal{D})$ for some value $y\sim \mathcal{P}_{Y_0}$. As such, an implicit density estimation mechanism $x_1,..., x_{|\mathcal{D}|}$ of $\mathcal{D}$ shifts the distribution $p(x_0|\mathcal{D})$ away from $\Pi_{X_0}$ and towards the regions of high probability under $\mathcal{P}_X^{\mathcal{D}}$. This implicit density estimation mechanism appears clearly in the limiting case where the aleatoric uncertainty increases since we observe that pdf $p(x_0|\mathcal{D})$ becomes $p(x_0|x_1,...,x_{|\mathcal{D}|})$. Conversely, when the aleatoric uncertainty decreases, this pdf is, under assumptions of identifibility and invertibility, $\pi_{X_0}(x_0)$. We will illustrate this effect in both contexts of regression (using the running example) and classification. 
As a consequence, the ppd in the discriminative approach indeed does not provide with an approximation of \eqref{posterior} with prior $\Pi_{X_0}$. It instead provides with an approximation of:
\begin{equation}\label{wrong_approx}
    \frac{p_X^{\mathcal{D}}(x_0)p_{Y|X}(y_0|x_0)}{\int p_X^{\mathcal{D}}(x)p_{Y|X}(y_0|x)\mathrm{d}x}.
\end{equation}
Consequently, a mismatch between the prior $\Pi_{X_0}$ and the distribution $\mathcal{P}_X^{\mathcal{D}}$, which effectively generated the $x_i$ values in $\mathcal{D}$, will result in a mismatch between the target posterior \eqref{posterior} and the ppd \eqref{posterior_predictive}. Subsequently, only the regions of space which are well represented by the $x_i$ values in dataset $\mathcal{D}$ will have high probability mass under the marginal $p(x_0|\mathcal{D})$, and hence, under the ppd $p(x_0|y_0, \mathcal{D})$. Though this argument relates $\mathcal{D}$ to the posterior $p(x_0|y_0,\mathcal{D})$ (via the distribution $p(x_0|\mathcal{D})$), we consider that this argument is not related to epistemic uncertainty as (i) the effect does not vanish when the number of recorded observation, i.e. the size of $\mathcal{D}$ increases; and (ii) the same effect can be observed when considering $p(x_0|y_0,\theta^*)$ where $\theta^*$ is a unique parameter (such as MLE or MAP) estimate.
\\
Finally, in the discriminative case, it is of particular interest to study the distribution $p(x_0|\mathcal{D})$ as it corresponds to the average prediction over observations $y_0$ since $\E_{y_0\sim \mathcal{P}_{Y_0}}\bracket{p(x_0|y_0,\mathcal{D})} \stackrel{D}{=} p(x_0|\mathcal{D})$.
% \begin{equation}
%     \E_{y_0\sim \mathcal{P}_{Y_0}}\bracket{p(x_0|y_0,\mathcal{D})} \stackrel{D}{=} p(x_0|\mathcal{D}).
% \end{equation}
This, together with the probability mass of $p(x_0|\mathcal{D})$ which favors the regions of $x$ values in $\mathcal{D}$, tells us that a discriminative model will favor the regions which are well represented in the dataset. In a classification task, the dominant labels will be predicted more often than the others, thus explaining that discriminative models indeed suffer from imbalanced dataset. We further emphasize this precise point using the illustrative running example, a classification example provided in supplementary materials (see section
\ref{classification_example}),
as well as in quantitative simulations in section \ref{simulations}.
    
\subsubsection*{\sl Illustrative running example}
We now leverage the affine regression example to illustrate the effects of the implicit prior on the ppd in the discriminative modeling approach. We first display an empirical approximation of the distribution $p(x_0|\mathcal{D})$. To that end, using equation \eqref{sampling_from_x0_given_D}, we obtain samples from this distribution via the two step sampling procedure $y\sim \mathcal{P}_{Y_0}$ and $x_0\sim p(x_0|y,\mathcal{D})$ (the second sampling step is detailed in the next paragraph \ref{gibbs_supervised}). Of course in practice, the first sampling step cannot be conducted as sampling from $p_{Y_0}(y)=\int p_{Y|X}(y|x)\pi_{X_0}(x)\mathrm{d}x$ requires sampling from the DGP which we recall is unknown by hypothesis, but in our example, we do resort to this sampling procedure for illustration purposes.

\begin{figure}[!ht]
    \centering
    \begin{subfigure}{0.6\textwidth}
    \centering
    \includegraphics[width=.8\linewidth]{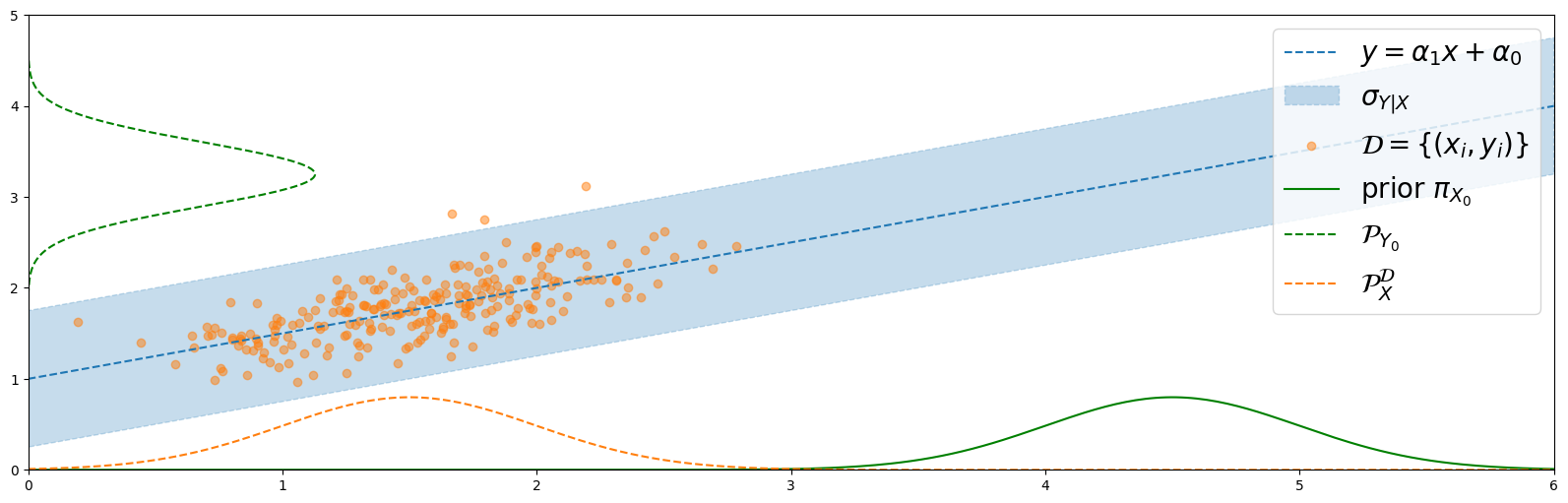}
    \centering
    \includegraphics[width=.8\linewidth]{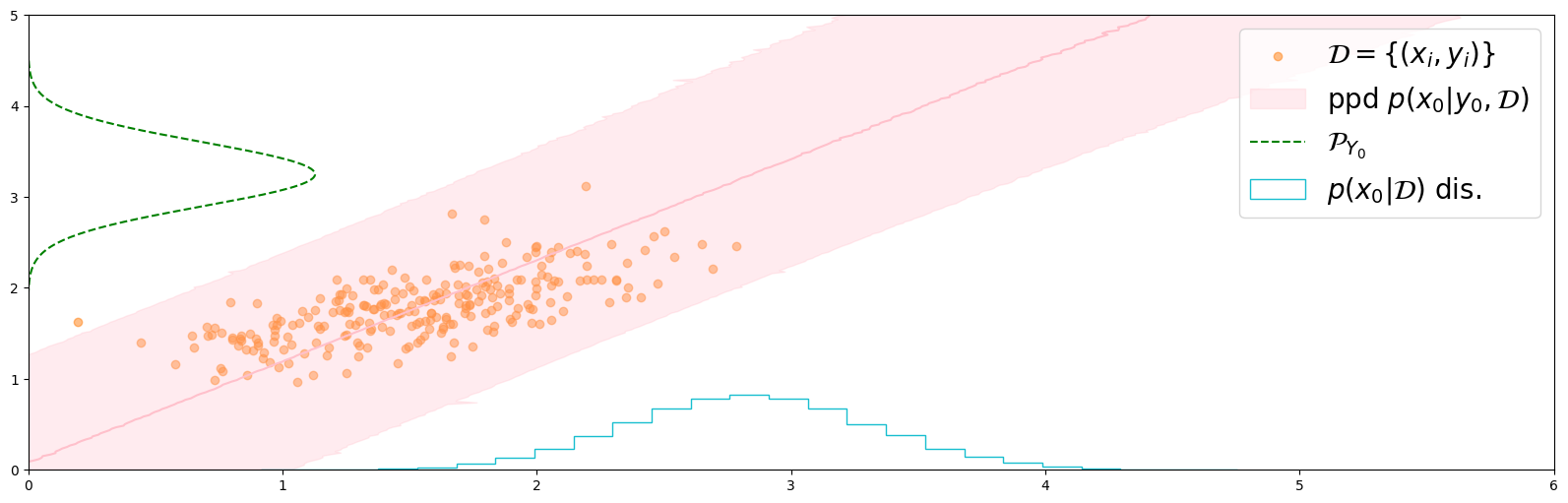}
    \end{subfigure}
    \begin{subfigure}{0.3\textwidth}
    \includegraphics[width=\linewidth]{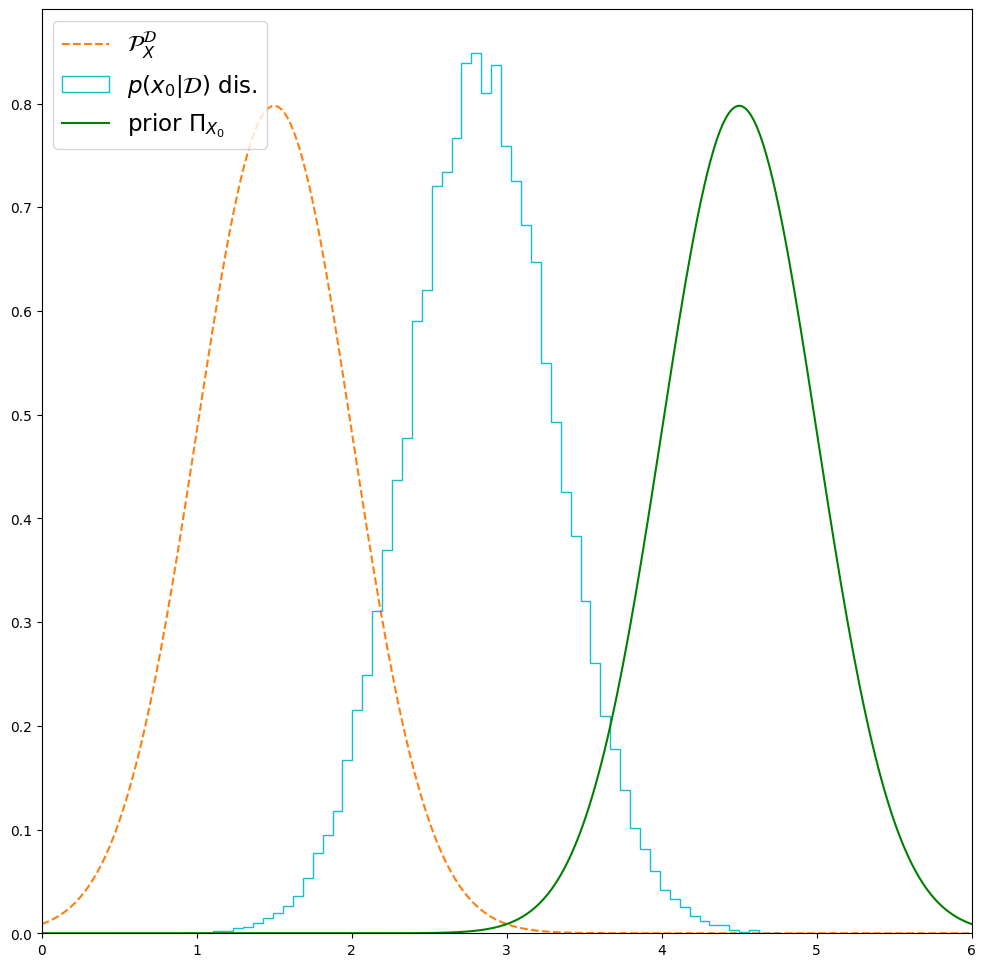}
    \end{subfigure}
    \caption{Empirical estimate of $p(x_0|\mathcal{D})$ in the discriminative setting: samples are obtained via $y\sim \mathcal{P}_{Y_0}$ (top) followed by $x_0\sim p(x_0|y,\mathcal{D})$ (bottom). This distribution corresponds to a trade-off between $\Pi_{X_0}$ and $\mathcal{P}_X^{\mathcal{D}}$ (right)}\label{fig:regression_x0_given_D}
\end{figure}

In figure \ref{fig:regression_x0_given_D}, an empirical estimate of $p(x_0|\mathcal{D})$ is obtained via the described two-step sampling procedure (upper-left and lower-left) and is plotted against prior $\Pi_{X_0}$ and $\mathcal{P}_{X}^{\mathcal{D}}$. We see that, in the discriminative setting, $p(x_0|\mathcal{D})$ (which we recall acts as a prior in \eqref{implicit_prior}) indeed corresponds to a trade-off between the two distributions and is shifted towards $\mathcal{P}_{X}^{\mathcal{D}}$ via an implicit density estimation mechanism from the $x$-values in $\mathcal{D}$. In this example, we can also visualize how the DGP affects the balance between $\Pi_{X_0}$ and $\mathcal{P}_{X}^{\mathcal{D}}$ which we now illustrate in the next figure \ref{fig:regression_x0_given_D_change_DGP}.
\begin{figure}[!ht]
    \centering
    \includegraphics[width=.8\linewidth]{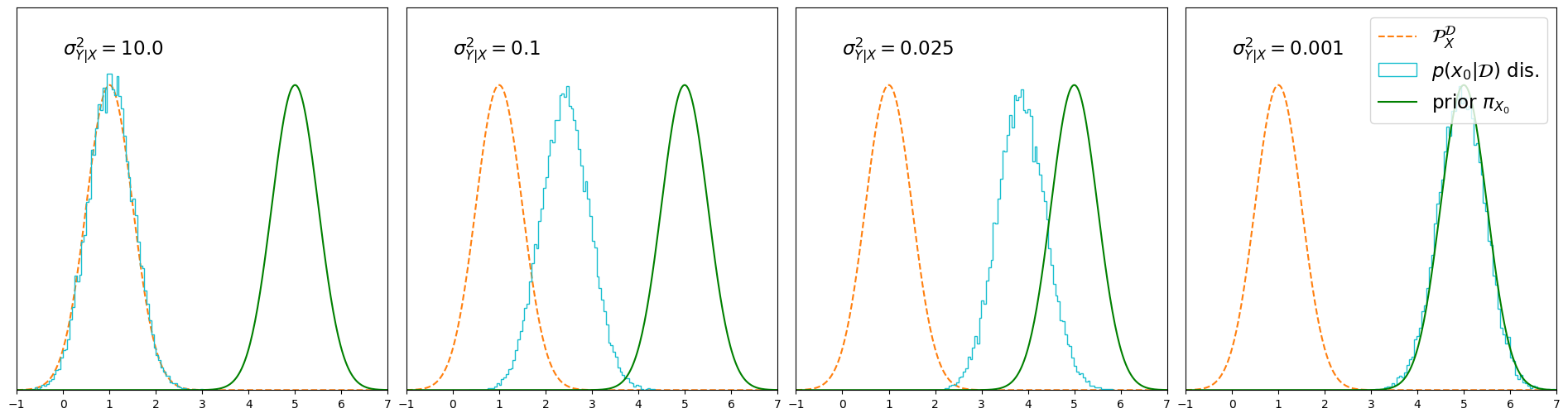}
    \caption{Varying degrees of aleatoric uncertainty in the DGP yield the distribution $p(x_0|\mathcal{D})$ to shift between $\Pi_{X_0}$ and $\mathcal{P}_X^{\mathcal{D}}$.}\label{fig:regression_x0_given_D_change_DGP}
\end{figure}

In this figure, on the one hand, we see that for lower values of $\sigma_{X|Y}$ (the noise standard deviation in the DGP \eqref{true_DGP_example}) the distribution $p(x_0|\mathcal{D})$ gets closer to $\Pi_{X_0}$; while, on the other hand, this distribution gets closer to $\mathcal{P}_{X}^\mathcal{D}$ for larger values of $\sigma_{X|Y}^2$. This example seems to hint that when the DGP is stained with high (resp. low) aleatoric uncertainty, the distribution $p(x_0|\mathcal{D})$ leans more towards $\mathcal{P}_X^{\mathcal{D}}$ (resp. $\Pi_{X_0}$). 
% In the next section \ref{classification_example} we will see that a similar effect can also be witnessed in a classification example. 
In section \ref{classification_example}, we provide with another example of classification and observe a similar effect.

\begin{figure}[!ht]
    \centering
    \includegraphics[width=.8\linewidth]{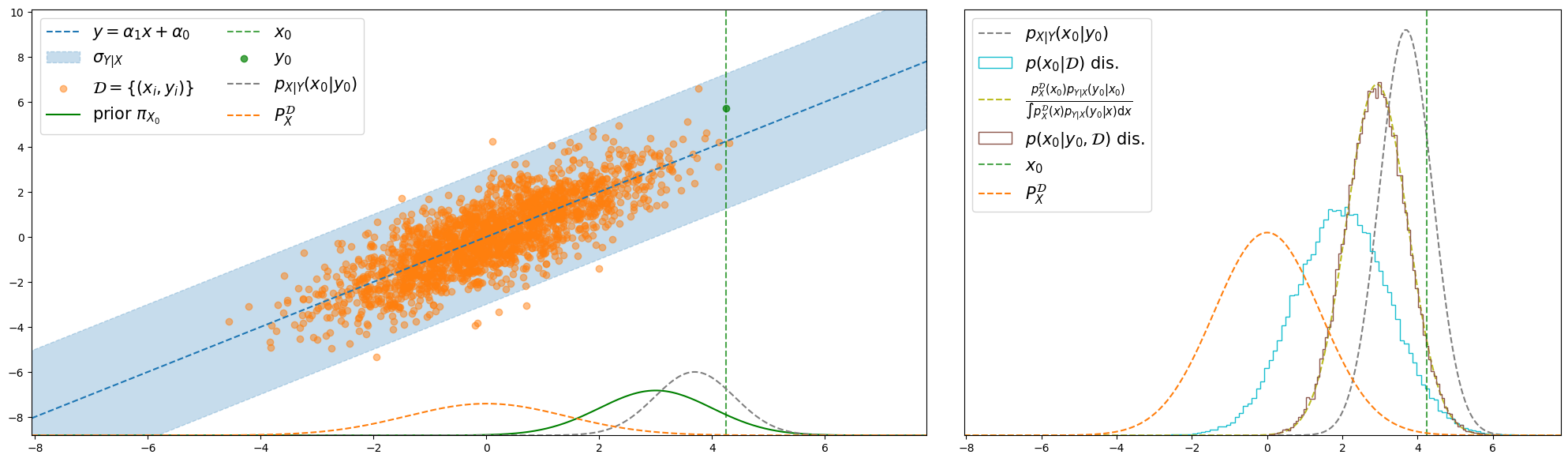}
    \centering
    \includegraphics[width=.8\linewidth]{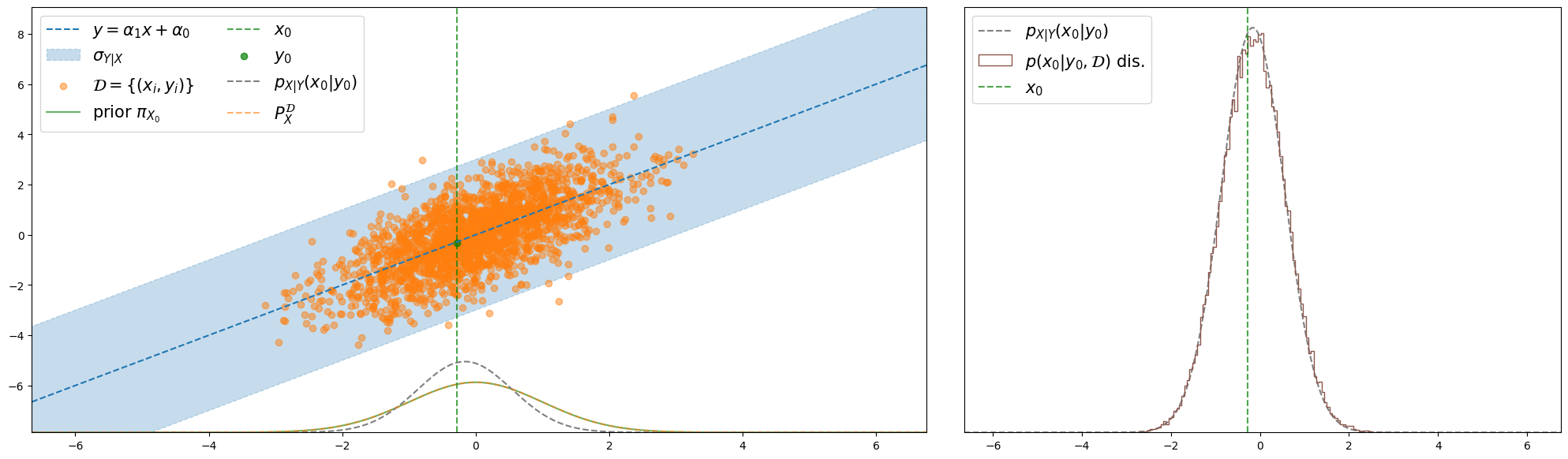}
    \caption{The $x_0$ marginal is inferred on $\mathcal{D}$: a mismatch between the prior $\Pi_{X_0}$ and $\mathcal{P}_X^{\mathcal{D}}$ results in a mislead approximation (first line). A discriminative approach is accurate in the case where these two distributions match (second line)}\label{fig:prior_dataset}
\end{figure}

As we have mentionned before, it therefore follows that the ppd in the discriminative case provides with an approximation of \eqref{wrong_approx}, which leads to a visible mismatch between the ppd and the true (unknown) posterior when the prior $\Pi_{X_0}$ and $\mathcal{P}_X^{\mathcal{D}}$ are different probability distributions, which we now illustrate. To that end, we compare this pdf to an histogram of samples from the ppd $p(x_0|y_0,\mathcal{D})$ with large $\mathcal{D}$ to remove the effect of epistemic uncertainty and we observe that they perfectly match. Conversely, when the prior $\Pi_{X_0}$ and $\mathcal{P}_X^{\mathcal{D}}$ are the same probability distribution then the ppd indeed approximates the true ppd. This is illustrated in figure \ref{fig:prior_dataset}.

\begin{figure}[!ht]
    \centering
    \includegraphics[width=.8\linewidth]{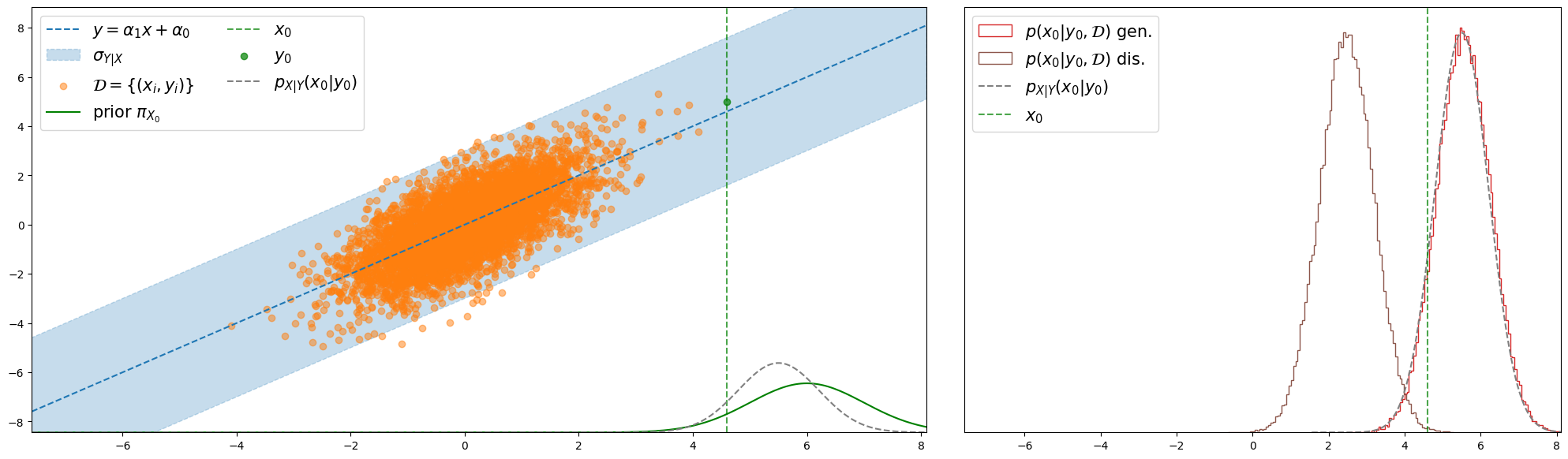}
    \caption{Illustration of inaccurate discriminative prediction when outside of the support of $\mathcal{D}$}\label{fig:prior_outside}
\end{figure}

We now illustrate a more problematic issue related to the same mechanism. Because the dataset shifts the distribution $p(x_0|\mathcal{D})$, which acts as a prior in the ppd, toward $\mathcal{P}_{X}^{\mathcal{D}}$ via an implicit density estimation mechanism of $\mathcal{P}_X^{\mathcal{D}}$ with $p(x_0|\mathcal{D})$ the ppd will only assign high probability to the regions of space which are assigned high probability under $p(x_0|\mathcal{D})$. The consequence of this in the case of affine modeling is that we are not able to predict accurately outside of the support induced by $\mathcal{D}$ as the ppd attributes little to no mass to the true value of $x_0$. A discriminative affine model cannot extrapolate to regions outside of the support of $\mathcal{D}$ and this conclusion argues, for once, in disfavor of a discriminative approach since an affine model, amongst all models, is expected to extrapolate well. Conversely, as a result of the explicit prior, the generative approach does not suffer from the same shortcoming and we observe that the affine generative model indeed produces a ppd which assigns high probability to the true value of $x_0$ and approximates the true unknown posterior. This is illustrated in figure \ref{fig:prior_outside}.

\subsubsection{\sl Classification example}\label{classification_example}

\begin{wrapfigure}{R}{5cm}
    \centering
    \includegraphics[width=.7\linewidth]{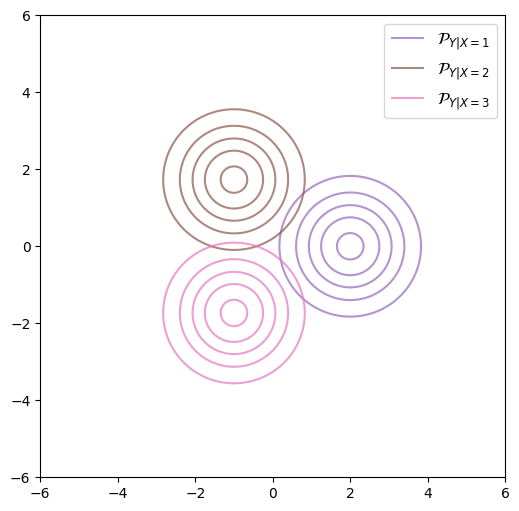}
    \caption{Contour plot of the DGP used in the classification example}\label{fig:dgp_classif}
\end{wrapfigure}

In the previous example of regression, we illustrated how a discriminative modeling approach builds an implicit prior via a density estimation mechanism, resulting in a poor approximation in the case of a mismatch between $\mathcal{P}_X^{\mathcal{D}}$ and the desired prior $\Pi_{X_0}$. We also illustrated how, in a discriminative model, the dataset $\mathcal{D}$, the prior $\Pi_{X_0}$ and the DGP interact to yield $p(x_0|\mathcal{D})$ which, via an implicit density estimation mechanism, is a trade-off between $\Pi_{X_0}$ and $\mathcal{P}_X^{\mathcal{D}}$. We now also illustrate this specific effect on a classification problem where $X$ is a Categorical variable taking value $c=1,...,C$. 

We consider a 2-dimensional example with $C=3$ where the DGP reads: $p_{Y|X=c}(y) = \mathcal{N}({\scriptstyle y; r_{Y|X}\bracket{ 
        \mathrm{Re}(\omega^c),\mathrm{Im}(\omega^c)}
         ^T, I_2})$  
with $\omega = e^{2i\pi/3}$. In this DGP, the aleatoric uncertainty can be controlled via the value of $r_{Y|X}$ which describes the distance of each Gaussian class to the origin ($r\_$ stands for radius). Indeed, the further the different classes are from each other, the easier it is to accurately classify a sample $y \sim \mathcal{P}_{Y_0}$ to its according unknown label.
The goal of this section is to illustrate the effect of the roles of distributions $\mathcal{P}_X^{\mathcal{D}}$ and $\Pi_{X_0}$ in the discriminative approach. We therefore select the two distributions $\mathcal{P}_X^{\mathcal{D}}$ from be different to one another: $\mathrm{Pr}(X=c) = (4-c)/6$ for $X\sim \Pi_{X_0}$ versus $\mathrm{Pr}(X=c) = c/6$  for $X\sim\mathcal{P}_X^{\mathcal{D}}$ for labels $c=1,2,3$. The distribution $\mathcal{P}_X^{\mathcal{D}}$ defines the frequencies of classes, and together with the DGP can produce a toy dataset $\mathcal{D}$ which is illustrated in the next figure \ref{fig:dataset_classif}. 

\begin{figure}[!ht]
    \includegraphics[width=.8\linewidth]{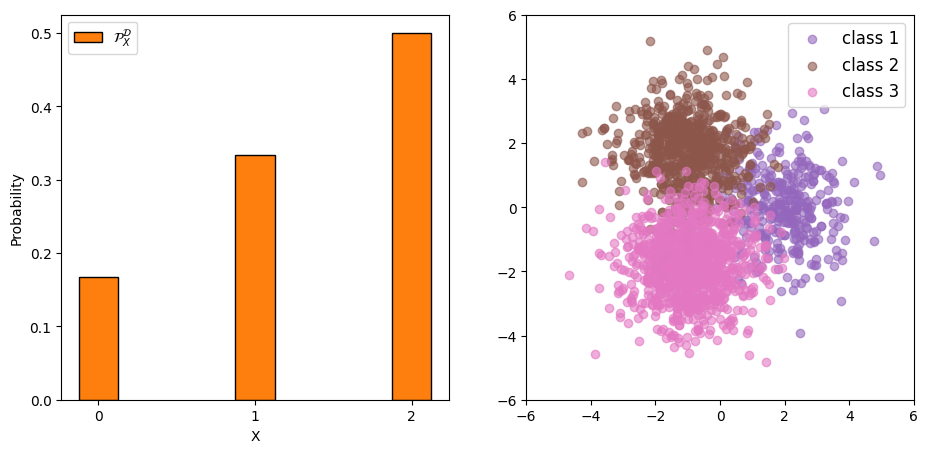}
    \caption{Dataset $\mathcal{D}$ with class distribution $\mathcal{P}_X^\mathcal{D}$}\label{fig:dataset_classif}
\end{figure}

As a discriminative model, we use a multinomial Logistic classifier which class probability reads:
\begin{equation}\label{logistic_classifier_prob}
    \mathrm{Pr}_\theta(X=c|Y) = \frac{\exp(\beta_{c}^TY + \beta_{c,0})}{\sum_{c=1}^C\exp(\beta_{c}^TY +\beta_{c,0})};
\end{equation}
where $\theta = \set{\beta_1,\beta_{1,0},...,\beta_C,\beta_{C,0}}$ with $\beta_c,\beta_{c,0} \in \mathbbm{R}^2 \times \mathbbm{R}$. Similarly to the continued regression example, we will display an empirical estimate of distribution $p(x_0|\mathcal{D})$ \eqref{x0_theta_D_discriminative} obtained by the sampling procedure: $y\sim \mathcal{P}_{Y_0}$ (via the unknown DGP which we use only for illustrative purposes), $\theta\sim p(\theta|\mathcal{D})$ and $x_0 \sim \mathrm{Categorical}(\mathrm{Pr}_\theta(X=1|Y),\mathrm{Pr}_\theta(X=2|Y),\mathrm{Pr}_\theta(X=3|Y))$ given by equation \eqref{logistic_classifier_prob}. 

A prior distribution over parameters $\Pi_\theta$ which would be conjugate to this logistic model would lead to a posterior $p(\theta|\mathcal{D})$ available in close form, but unfortunately, such a conjugate prior does not exists amongst the usual probability distributions. We therefore use a simple Gaussian prior over parameter $\theta$ and resort to sampling from the ppd using an Metropolis-Hastings MCMC algorithm, though a Gibbs sampling scheme is also available in this setting \cite{holmes2006bayesian}. In the next figure, we display empirical estimates of $p(x_0|\mathcal{D})$ against $\Pi_{X_0}$ and $\mathcal{P}_{X}^{\mathcal{D}}$. 
\begin{figure}[!ht]
    \centering
    \includegraphics[width=.8\linewidth]{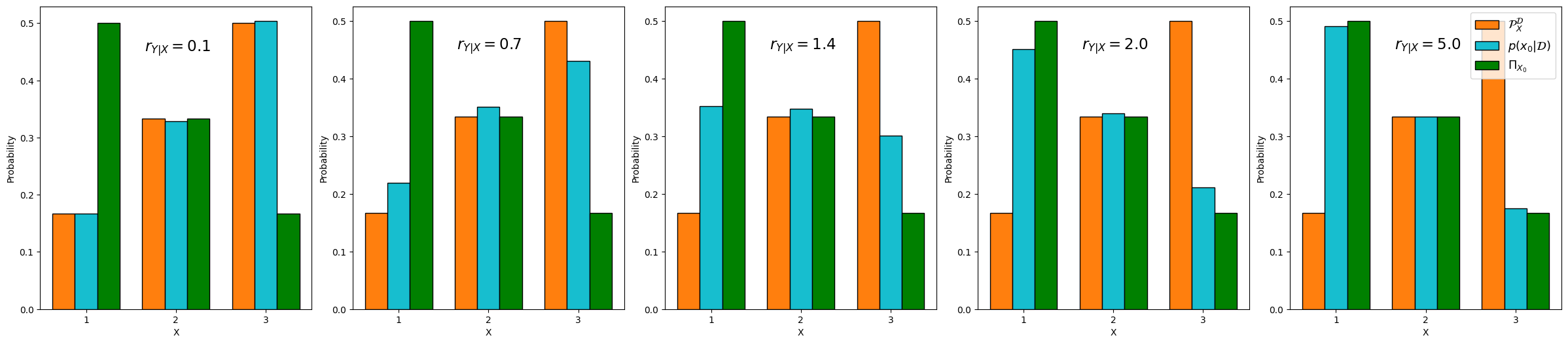}
    \caption{Varying degrees of aleatoric uncertainty in the DGP yield the distribution $p(x_0|\mathcal{D})$ to shift between $\Pi_{X_0}$ and $\mathcal{P}_X^{\mathcal{D}}$.}\label{fig:classification_x0_given_D_change_DGP}
\end{figure}
 This figure once again illustrates that distribution $p(x_0|\mathcal{D})$ (which we recall acts as a prior in the ppd \eqref{implicit_prior}) corresponds to a trade-off between $\Pi_{X_0}$ and $\mathcal{P}_X^{\mathcal{D}}$. With a very similar interpretation to that of figure \ref{fig:regression_x0_given_D_change_DGP}, this figure also seems to hint that the dynamic of the DGP dictates the trade-off between the two distributions: high (resp. low) aleatoric uncertainty shifts $p(x_0|\mathcal{D})$ more towards $\mathcal{P}_{X}^{\mathcal{D}}$ (resp. $\Pi_{X_0}$).

\subsection{Gibbs sampling from the ppd}\label{gibbs_supervised}
We now come to sampling from the ppd. Samples from the ppd can be obtained in two distinct ways. If exact computation of expectation w.r.t. $p(\theta|\mathcal{D})$ is feasible, then \eqref{generative_posterior_predictive} and \eqref{discriminative_posterior_predictive} give tractable expressions (possibly up to a constant) for the ppd. However, the integrals in these equations admit closed form expressions only in specific cases when using conjugate models, and exact integration is, more often than not, unfeasible making the posterior pdf intractable. So in practice, we rather resort to sampling via the joint distribution since, as we explained before, sampling from the joint distribution \eqref{joint} produce samples $x_0$ that are distributed according to the ppd \eqref{posterior_predictive}. 

With that regard, in both modeling cases, the joint pdf, \eqref{generative_joint} and \eqref{discriminative_joint}, can be computed (at least up to a constant), and so we can sample from the joint distribution via the pdf. More conveniently, the discriminative approach yields a specific factorization of its joint pdf \eqref{discriminative_joint} which enables a sequential sampling procedure with $\theta \sim p(\theta|\mathcal{D})$ and $x_0\sim p_\theta(x_0|y_0)$. It is therefore motivated to construct models for which the posterior probability distribution of parameters  By contrast, the generative approach does not induce the same factorization and does not benefit from the same convenient sequential sampling scheme. In this section, we propose a general scheme for sampling from the ppd which can be applied to both modeling approach. 

We now propose a scheme that enables to sample from the joint distribution \eqref{joint}, and therefore from its $x_0$ marginal which is nothing but the ppd \eqref{posterior_predictive}. This sampling scheme is based on the notorious Gibbs sampling \cite{casella1992explaining}\cite{gelfand2000gibbs}
which is an MCMC algorithm \cite{roberts2004general}
that applies specifically to a joint distribution $p(u,v)$ with a markov transition $q_t(u^{(t+1)}, v^{(t+1)}|u^{(t)}, v^{(t)}) = p(u^{(t+1)}|v^{(t+1)})p(v^{(t+1)}|u^{(t)})$. This transition leaves the joint distribution $p(u,v)$ invariant since the Gibbs algorithm can be seen as two steps of Metropolis Hastings \cite{chib1995understanding} transition where the acceptance probability is $1$. We apply the principle of Gibbs sampling to the joint distribution $p(x_0,\theta|y_0,\mathcal{D})$ in the generative (resp. discriminative) setting. First, conditionally on the current model $\theta^{(t-1)}$, $x_0$ is distributed according to the posterior for that model. So the first step of the markov transition is to draw $x_0^{(t)}$ from equation \eqref{generative_posterior} (resp. \eqref{discriminative_posterior}) with $\theta^{(t-1)}$. Then, conditionally on the current value of $x_0^{(t)}$, $\theta$ is distributed according to $p(\theta|x_0^{(t)},y_0,\mathcal{D})$ and so the second step of the markov transition consists in sampling $\theta^{(t)}\sim p(\theta|\mathcal{D}_+^{(t)})$ with the analogous of \eqref{generative_model_posterior} (resp. \eqref{discriminative_model_posterior}) where $\mathcal{D}_+^{(t)} = \mathcal{D} \cup \{(x_0^{(t)},y_0)\}$. 
We summarize this Gibbs sampler in algorithm \ref{alg:gibbs_semi_supervised} (for the moment readers should disregard the red parts of the algorithm, as they are related to the semi-supervised learning task covered in section \ref{semi-supervised}).
% \textcolor{blue}{removed here}
% We summarize this Gibbs sampling in algorithm \ref{alg:gibbs_supervised}. 

% \textcolor{blue}{removed here}
% \begin{algorithm}
% \caption{Gibbs sampling from $p(x_0|y_0,\mathcal{D})$ in the generative (resp. discriminative) setting}
% \label{alg:gibbs_supervised}
% \begin{algorithmic}
% \REQUIRE{$ y_0,\mathcal{D}$, number of steps $T$}
% \STATE{$\theta^{(0)}\sim p(\theta|\mathcal{D})$}
% \FOR{$t=1$ to  $T$}
% \STATE{$x_0^{(t)} \sim p(x_0|y_0,\theta^{(t-1)})$ with \eqref{generative_posterior} (resp. \eqref{discriminative_posterior}), set $\mathcal{D}_+^{(t)} =  \mathcal{D}\cup\{(x_0^{(t)},y_0)\}$}
% \STATE{$\theta^{(t)} \sim p(\theta|\mathcal{D}_+^{(t)})$ with \eqref{generative_model_posterior} (resp. \eqref{discriminative_model_posterior})}
% \ENDFOR
% \RETURN $x_0^{(T)}$
% \end{algorithmic}
% \end{algorithm}

Though this algorithm is written in a similar fashion in both modeling approaches, the conclusions with regard to the different behaviours of the two modeling approaches presented in the previous sections still hold as they are the result of a structural difference between the generative and discriminative approach. This algorithm will be especially useful in the semi-supervised setting, which we describe in the next section \ref{semi-supervised}.

In the case of multiple observations $y_{0,1},...,y_{0,N_0}$ as in section \ref{handling_mulitple_observations}, the previous algorithm can be effortlessly adjusted in the generative case (recall that the discriminative case is not compatible with multiple observations, see section \ref{handling_mulitple_observations}). Indeed in this setting, at time $t$, we first sample $x_0^{(t)}$ according to \eqref{generative_posterior_multiple} for current model $\theta^{(t-1)}$; and then the dataset $\mathcal{D}$ is augmented into $\mathcal{D}_+^{(t)} =  \mathcal{D}\bigcup_{i=1}^{N_0}\{(x_0^{(t)},y_{0,i})\}$.

\subsubsection*{\sl Illustrative running example}
We now come back to the continued example of affine modeling to provide an example of the Gibbs algorithm mechanism. We assume prior knowledge over parameter $\theta$ in the form of $\pi_\Theta(\theta) = \mathcal{N}(\beta; \mu_\beta, \Sigma_\beta)\mathrm{I}\Gamma(\sigma^2; \lambda, \eta)$ where $\mu_\beta \in \mathbb{R}^2$ and $\Sigma_\beta \in \mathbb{R}^{2\times 2}$ is a covariance matrix, $\mathrm{I}\Gamma$ is the pdf associated with an inverse Gamma distribution and $\lambda,\eta >0$ are respectively the shape and scale parameters of the corresponding Gamma distribution. Unfortunately, the posterior $p(\theta|\mathcal{D})$ does not admit a closed form expression; but, at least, this choice of conjugate priors allows for both conditionals to be tractable. We first explicit these two conditionals pdf in the generative (resp. discriminative) setting: 

\begin{align}
    &\label{affine_coefficients_posterior_generative}p(\beta|\sigma^2,\mathcal{D}) \stackrel{G}{=} \mathcal{N}(\beta;(\frac{\mathbf{X}^T\mathbf{X}}{\sigma^2} + \Sigma_{\beta}^{-1})^{-1}(\frac{\mathbf{X}^T\mathbf{Y}}{\sigma^2}+ \Sigma_{\beta} \mu_{\beta}),
    (\frac{\mathbf{X}^T\mathbf{X}}{\sigma^2}+ \Sigma_{\beta}^{-1})^{-1}), \\&\nonumber\text{ where } \mathbf{X} \stackrel{G}{=} \begin{bmatrix}x_1, 1\\ ... \\x_{|\mathcal{D}|},1\end{bmatrix}, \mathbf{Y} \stackrel{G}{=} \begin{bmatrix}y_1\\ ... \\y_{|\mathcal{D}|}\end{bmatrix};\\
    &p(\sigma^2|\beta, \mathcal{D}) \stackrel{G}{=} \mathrm{I}\Gamma(\sigma^2; \lambda + \frac{|\mathcal{D}|}{2}, \eta + \sum_{i=1}^{|\mathcal{D}|}\frac{(y_i - \beta_1x_i - \beta_0)^2}{2}).\label{noise_variance_posterior_generative}
\end{align}
\begin{align}
    &\label{affine_coefficients_posterior_discriminative}p(\beta|\sigma^2,\mathcal{D}) \stackrel{D}{=} \mathcal{N}(\beta;(\frac{\mathbf{Y}^T\mathbf{Y}}{\sigma^2} + \Sigma_\beta^{-1})^{-1}(\frac{\mathbf{Y}^T\mathbf{X}}{\sigma^2} + \Sigma_\beta \mu_\beta),
    (\frac{\mathbf{Y}^T\mathbf{Y}}{\sigma^2} + \Sigma_\beta^{-1})^{-1}),
      \\&\nonumber\text{ where } \mathbf{Y} \stackrel{D}{=} \begin{bmatrix}y_1, 1\\ ... \\y_{|\mathcal{D}|},1\end{bmatrix}, \mathbf{X} \stackrel{D}{=} \begin{bmatrix}x_1\\ ... \\x_{|\mathcal{D}|}\end{bmatrix};
    \\
    &p(\sigma^2|\beta, \mathcal{D}) \stackrel{D}{=} \mathrm{I}\Gamma(\sigma^2; \lambda + \frac{|\mathcal{D}|}{2}, \eta + \sum_{i=1}^{|\mathcal{D}|}\frac{(x_i - \beta_1y_i - \beta_0)^2}{2}).\label{noise_variance_posterior_discriminative}
\end{align}

So in both modeling cases, the posterior distribution $p(\theta|\mathcal{D})$ can be sampled from using a Gibbs scheme by sequentially sampling these two conditionnals. Then, from a Gibbs sampling of affine models, we can almost effortlessly obtain samples from the ppd  by including the additional step of sampling $x_0$ from $p(x_0|y_0,\theta)$ for the current model parameters within the Gibbs sequential Markovian transition. 
Again, in supplementary material, we summarize this Gibbs sampler in the specific case of affine homoskedastic modelling,
see algorithm  \ref{alg:gibbs_affine_semi_supervised} 
(readers should disegard the steps in red for now, as they are related to semi-supervised learning which we now discuss). 

% \textcolor{blue}{removed here}
% \begin{algorithm}
% \caption{Gibbs sampling from $p(x_0|y_0,\mathcal{D})$ using a generative (resp. discriminative) homoskedastic affine model}
% \label{alg:gibbs_affine_supervised}
% \begin{algorithmic}
% \REQUIRE{$ y_0,\mathcal{D}, \mathcal{Y}$, number of steps $T$}
% \STATE{${\sigma^{2}}^{(0)} \sim \Gamma^{-1}(\sigma^2;\lambda, \eta)$}
% \STATE{$\beta^{(0)}\sim p(\beta|{\sigma^2}^{(0)},\mathcal{D})$ with \eqref{affine_coefficients_posterior_generative} (resp. \eqref{affine_coefficients_posterior_discriminative})}
% \FOR{$t=1$ to  $T$}
% \STATE{$x_0^{(t)} \sim p(x_0|y_0,\beta^{(t-1)}, {\sigma^2}^{(t-1)})$ with \eqref{x0_posterior_affine_generative} (resp. \eqref{affine_discriminative_model}), set $\mathcal{D}_+^{(t)} = \mathcal{D} \cup \{(x_0^{(t)},y_0)\}$}
% \STATE{${\sigma^2}^{(t)} \sim p(\sigma^2|\beta^{(t-1)},\mathcal{D}_+^{(t)})$ with \eqref{noise_variance_posterior_generative} (resp. \eqref{noise_variance_posterior_discriminative})}
% \STATE{$\beta^{(t)} \sim p(\beta|{\sigma^2}^{(t)},\mathcal{D}_+^{(t)})$ with \eqref{affine_coefficients_posterior_generative} (resp. \eqref{affine_coefficients_posterior_discriminative})}
% \ENDFOR
% \RETURN $x_0^{(T)}$
% \end{algorithmic}
% \end{algorithm}

\section{Bayesian Semi-Supervised learning}\label{semi-supervised}
In this section, we now build upon the equations, arguments and conclusions presented in the previous sections to tackle the problem of semi-supervised learning. As we have mentionned before, supervised learning techniques use the observed variables and their corresponding labels to build a model which capture the dependency between two rv and which can be used to make predictions about the label conditionally on the value of observed rv. 
Conversely, unsupervised learning \cite{hinton1999unsupervised}\cite{hastie2009unsupervised} take interest in learning pattern in a data distribution without considering the notion of associated labels. Structure can be represented by data clusters obtained by K-means \cite{lloyd1982least}\cite{likas2003global}, graph-based clustering methods such as spectral clustering \cite{ng2001spectral} or Louvain method \cite{blondel2008fast}, or via maximum-likelihood in a mixture probability distribution model \cite{dempster1977maximum}. In the beginning of this paper, we explained that the Generative approach for modeling the unknown posterior, should not be confused with the tasks and techniques of Generative modeling. These techniques can also be considered as unsupervised learning as they enable to capture the structure from univariate data such that the corresponding probabilistic model can be sampled from easily in order to obtain observations which are approximately distributed according to the dataset distribution. Most popular methods include Variational AutoEncoders \cite{kingma2013auto}, Generative Adversarial Networks \cite{goodfellow2020generative}, Normalizing Flows \cite{papamakarios2021normalizing} and Diffusion models \cite{sohl2015deep} but this is beyond the scope of this paper. Semi-supervised learning does however lie within the scope of this paper as it aims to obtain a conditional model to predict label from observations, but the goal is to infer the model from both a labeled dataset and unlabeled observations. In this section we now build upon the previous arguments and discuss the compatibility of both learning approach with this learning task. 

\subsection{The learning task}
In section \ref{supervised} and onward, we presented the general task of learning a model for the posterior \eqref{posterior} using a set of \emph{labeled} observations $\mathcal{D}$, and how to predict about an $x_0$ given a corresponding observation $y_0$ with the ppd, which we now refer to as a supervised learning task. However, in many statistical learning settings, we also dispose of \emph{unlabeled} observations. They corresponds to values $\widetilde{y}_j$, which we know (or assume) are produced by the DGP, but for an unknown values $\widetilde{x}_j$ for which we assume prior knowledge $\Pi_{\widetilde{X}_j}$: $\mathcal{Y} = \{\widetilde{y}_j| \exists \widetilde{x_j} \sim \Pi_{\widetilde{X}_j}, \widetilde{y}_j \sim \mathcal{P}_{Y|X}(Y|X = \widetilde{x_j}) \}_{j=1}^{|\mathcal{Y}|}$. 
% \begin{equation}
%     \mathcal{Y} = \{\widetilde{y}_j| \exists \widetilde{x_j} \sim \Pi_{\widetilde{X}_j}, \widetilde{y}_j \sim \mathcal{P}_{Y|X}(Y|X = \widetilde{x_j}) \}_{j=1}^{|\mathcal{Y}|}.
% \end{equation}
When the observations in $\mathcal{D}$ and $\mathcal{Y}$ cover different regions of the observation space, and/or when $\mathcal{Y}$ has a significant amount of elements, then the unlabeled observations $\mathcal{Y}$ may contain significant or non-negligible information \cite{nigam2000text}. In this context, a semi-supervised learning task aims at inferring a model from both labeled and unlabeled observations. This question has risen in importance in importance where we dispose of a lot of unlabeled observations, but where the labelling tasks is expensive (as it is the case when the labelling needs to be conducted by a human operator).

\begin{figure}[!ht]
\centering
\includegraphics[width=.7\textwidth]{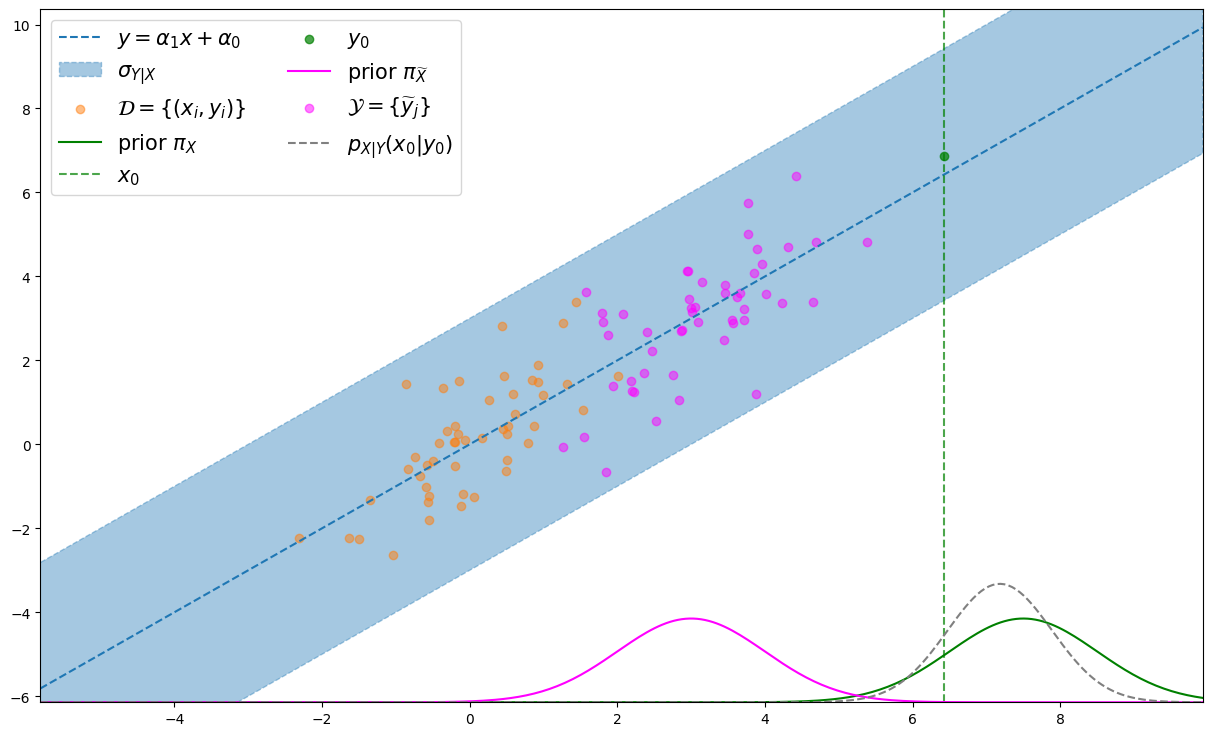}
\caption{Affine semi-supervised regression setting}\label{fig:semi_supervised_affine_setting}
\end{figure}
The ppd \eqref{posterior_predictive} then becomes:
\begin{equation}\label{posterior_predictive_semi_supervised}
    p(x_0|y_0,\mathcal{D}, \mathcal{Y}) = \int_\Theta p(x_0|y_0, \theta) p(\theta|y_0,\mathcal{D}, \mathcal{Y})\mathrm{d}\theta,
\end{equation}
and we aim to compute this pdf, or sample this distribution if exact computation of the integral is not feasible. This formulation is more general than the one described in section \ref{supervised} and it reduces to supervised learning in the case where $\mathcal{Y} = \varnothing$.

Throughout this section, we will carry on using the continued example of affine modeling to illustrate the arguments and conclusions. To that end, we suppose that, in addition to $\mathcal{D}$, we also dispose of unlabeled observations $\widetilde{y}_j$ produced from the DGP \eqref{true_DGP_example} via an unknown label $\widetilde{x}_j$ for which we suppose prior knowledge in the form of a prior $\Pi_{\widetilde{X}}$ which is supposed to be the same for all $j=1,...,|\mathcal{Y}|$ and which we consider to be Gaussian $\pi_{\widetilde{X}}(\widetilde{x}_j) = \mathcal{N}(\widetilde{x}_j; \mu_{\widetilde{X}}, \sigma^2_{\widetilde{X}})$. 
The semi-supervised setting is illustrated in figure \ref{fig:semi_supervised_affine_setting}.

\subsection{Both modeling confronted to the semi-supervised learning task}\label{semi supervised with gen vs dis}
We now confront the two modeling approaches to the specific problem of semi-supervised learning by analysing the model posterior which reads:
\begin{equation}\label{model_posterior_semi_supervised}
    p(\theta|y_0,\mathcal{D}, \mathcal{Y}) = p(\theta|\mathcal{D})\frac{p(y_0|\theta)p(\mathcal{Y}|\theta)}{p(y_0, \mathcal{Y}|\mathcal{D})}.
\end{equation}
We first explain that the discriminative approach does not allow for Bayesian semi-supervised learning. To see this, recall the conclusion of section \ref{with_without_observation}: when we do not know $x_0$, the posterior over $\theta$ does not depend on $y_0$ and so this observation does not carry any information to the discriminative models. Therefore, the same applies for the elements of $\mathcal{Y}$: since we do not know the label $\widetilde{x}_j$, the unlabeled observation does not bring any information on $\theta$ as the posterior over models does not depend on $\widetilde{y}_j$. Finally, the model posterior \eqref{model_posterior_semi_supervised} reduces to $p(\theta|\mathcal{D})$, \eqref{posterior_predictive_semi_supervised} reduces to \eqref{discriminative_posterior_predictive} and all the other equations in section concerning the discriminative modeling approach remain unchanged. 

Conversely, the generative approach indeed allows for semi-supervised learning. In section \ref{with_without_observation}, we explained that, even though we do not know the value of $x_0$, the posterior distribution over models still depends on the observation $y_0$ indirectly through the prior $\Pi_{X_0}$. With a similar argument, we understand that the unlabeled data $\mathcal{Y}$ indeed carry information on model $\theta$. We write: $ p(\mathcal{Y}|\theta) \stackrel{G}{=} \prod_{j=1}^{|\mathcal{Y}|}\int p_\theta(\widetilde{y}_j|\widetilde{x}_j) \pi_{\widetilde{X}_j}(\widetilde{x}_j)\mathrm{d}\widetilde{x}_j$.
% \begin{equation}\label{likelihood_unlabeled_observations}
%      p(\mathcal{Y}|\theta) \stackrel{G}{=} \prod_{j=1}^{|\mathcal{Y}|}\int p_\theta(\widetilde{y}_j|\widetilde{x}_j) \pi_{\widetilde{X}_j}(\widetilde{x}_j)\mathrm{d}\widetilde{x}_j
% \end{equation}
So, even though we do not know the label $\widetilde{x}_j$, probable generative models $\theta$ under \eqref{model_posterior_semi_supervised} produce, with high probability, the value of $\widetilde{y}_j$ for some unknown label distributed under the prior $\Pi_{\widetilde{X}_j}$. In a classification setting, this term can indeed be computed as a tractable finite sum \cite{atanov2019semi}\cite{izmailov2020semi}, but in general, this term is only available in integral form making the joint pdf $p(x_0,\theta|y_0,\mathcal{D}, \mathcal{Y}) \stackrel{G}{\propto}\pi_{X_0}(x_0)p_\theta(y_0|x_0)p(\theta|\mathcal{D})p(\mathcal{Y}|\theta)$,
% \begin{equation}
%     p(x_0,\theta|y_0,\mathcal{D}, \mathcal{Y}) \stackrel{G}{\propto}\pi_{X_0}(x_0)p_\theta(y_0|x_0)p(\theta|\mathcal{D})p(\mathcal{Y}|\theta) \label{generative_joint_semi_supervised},\\
% \end{equation}
intractable, not even up to a constant. This raises the question of sampling from the ppd since its pdf is intractable. In the next section, we propose to use a variation of the Gibbs algorithm presented in section \ref{gibbs_supervised}, and which allows to sample from this ppd.

\subsection{A Gibbs sampling algorithm for semi-supervised learning}\label{gibbs_semi_supervised}
In this section we extend the previous Gibbs sampling algorithm presented in section \ref{gibbs_supervised} for sampling from the ppd \eqref{posterior_predictive_semi_supervised} in the case of generative semi-supervised learning.
To that end, we apply the Gibbs mechanism of sequentially sampling the conditional distributions in the joint distribution $p(x_0, \widetilde{x}_1,...,\widetilde{x}_{|\mathcal{Y}|}, \theta|y_0, \mathcal{D}, \mathcal{Y})$. Firstly, conditionally on the current value of $\theta^{(t-1)}$, the labels $x_0,\widetilde{x}_1,...,\widetilde{x}_{|\mathcal{Y}|}$ are independent and each distributed according to its own posterior distribution. So the first step of the Gibbs markovian transition is to sample $x_0^{(t)}\sim p(x_0|y_0,\theta^{(t-1)})$ with equation \eqref{generative_posterior} (as in section \ref{gibbs_supervised}) and $\widetilde{x}_j^{(t)} \sim p(\widetilde{x}_j|\widetilde{y}_j,\theta^{(t-1)})$. Secondly, conditionally on the current label values $x_0^{(t)}, \widetilde{x}_1^{(t)},...,\widetilde{x}_{|\mathcal{Y}|}^{(t)}$, the model parameters are distributed according to $p(\theta|\mathcal{D},x_0^{(t)}, y_0, \widetilde{x}_1^{(t)}, 
\widetilde{y}_1,...,\widetilde{x}_{|\mathcal{Y}|}^{(t)},\widetilde{y}_{|\mathcal{Y}|})$. So the second step of the Gibbs markovian transition is to sample $\theta^{(t)} \sim p(\theta|\mathcal{D}_+^{(t)})$ analogous of equation \eqref{generative_model_posterior} where $\mathcal{D}_+^{(t)} = \mathcal{D} \cup(x_0^{(t)},y_0)\cup\{(\widetilde{x}^{(t)}_j, \widetilde{y}_j)\}_{j=1}^{|\mathcal{Y}|} $. We summarize this Gibbs mechanism in algorithm \ref{alg:gibbs_semi_supervised} and we highlight in red the steps which are effectively responsible for semi-supervised learning.

\begin{algorithm}
\caption{Gibbs sampling from $p(x_0|y_0,\mathcal{D}, \textcolor{red}{\mathcal{Y}})$ in the generative (resp. discriminative) setting}
\label{alg:gibbs_semi_supervised}
\begin{algorithmic}
\REQUIRE{$ y_0,\mathcal{D}, \textcolor{red}{\mathcal{Y}}$, number of steps $T$}
\STATE{$\theta^{(0)}\sim p(\theta|\mathcal{D})$}
\FOR{$t=1$ to  $T$}
\STATE{$x_0^{(t)} \sim p(x_0|y_0,\theta^{(t-1)})$ with \eqref{generative_posterior} (resp. \eqref{discriminative_posterior}}
\textcolor{red}{\FOR{all $\widetilde{y}_j \in \mathcal{Y}$}
\STATE{sample $\widetilde{x}_j^{(t)} \sim p(\widetilde{x}_j|\widetilde{y}_j,\theta^{(t-1)}) = \frac{p_{\theta^{(t-1)}}(\widetilde{y}_j|\widetilde{x}_j)\pi_{\widetilde{X}_j}(\widetilde{x}_j)}{p(\widetilde{y}_j|\theta^{(t-1)})}$}
\ENDFOR}
\STATE{set $\mathcal{D}_+^{(t)} = \mathcal{D} \cup (x_0^{(t)},y_0)$\textcolor{red}{$\cup\{(\widetilde{x}^{(t)}_j, \widetilde{y}_j)\}_{j=1}^{|\mathcal{Y}|}$} and $\theta^{(t)} \sim p(\theta|\mathcal{D}_+^{(t)})$ with \eqref{generative_model_posterior} (resp. \eqref{discriminative_model_posterior})}
\ENDFOR
\RETURN $x_0^{(T)}$
\end{algorithmic}
\end{algorithm}

In the case of semi-supervised learning setting, this Gibbs algorithm is all the more crucial. Indeed, while in the supervised context the Gibbs approach was only an alternative option to sampling from the joint distribution which pdf \eqref{generative_joint} could be computed up to a constant; in the case of semi-supervised learning however, it is possible that this joint pdf cannot be evaluated, not even up to a constant and the Gibbs approach is therefore very convenient for sampling the corresponding ppd. 

We have written the Gibbs algorithm with including $\mathcal{Y}$ to perform semi-supervised learning for both modeling approach but of course, as we have mentionned before, the semi-supervised ppd \eqref{posterior_predictive_semi_supervised} reduces to the supervised ppd \eqref{posterior_predictive} in the discriminative setting, so this Gibbs algorithm, even though it involves $\mathcal{Y}$, is not able to leverage any information from the unlabeled observations in this modeling approach. We therefore would like to stress that the semi-supervised learning is not enabled by the Gibbs procedure itself but rather by using a generative modeling instead of a discriminative one which induces different conditional dependency between all the rv.
We proposed the Gibbs algorithm as a way to sample the joint distributions $p(x_0,\theta|y_0,\mathcal{D},\mathcal{Y})$ in the case of generative modeling (with possibly $\mathcal{D}=\varnothing$ in the supervised setting) which is particularly convenient to use because of the conditional independence (w.r.t. $\theta$) of labels $x_0, \widetilde{x}_1,...,\widetilde{x}_{|\mathcal{Y}|}$. However, this Gibbs scheme it is only a possible approach for sampling from the corresponding ppd which effectively depends on $\mathcal{Y}$.

We now come back to the continued example of affine modeling and use it to illustrate the practical use of the Gibbs sampling algorithm which we use to illustrate, respectively, the compatibility and incompatibility between the generative and discriminative approaches and the semi-supervised learning. In algorithm
% broken reference \ref{alg:gibbs_affine_semi_supervised} 
SM3.1
we first explicit how to build upon the supervised Gibbs sampling algorithm of the supervised ppd 
% \textcolor{blue}{removed here}
% \eqref{alg:gibbs_affine_supervised}
presented in section \ref{gibbs_supervised}
to obtain a Gibbs sampling scheme of the semi-supervised ppd.

In this case, the generative and discriminative approach yield two different equations for the posterior $p(\widetilde{x}_j|\widetilde{y}_j,\theta)$:
\begin{align}
    &p(\widetilde{x}_j|\widetilde{y}_j,\beta, \sigma^2) \stackrel{G}{=} \mathcal{N}(\widetilde{x}_j; (\frac{1}{\sigma^2} + \frac{1}{\sigma_{\widetilde{X}}^2})^{-1}(\frac{\beta_1(\widetilde{y}_j - \beta_0)}{\sigma^2} + \frac{\mu_{\widetilde{X}}}{\sigma_{\widetilde{X}}^2}),
    (\frac{1}{\sigma^2} + \frac{1}{\sigma_{\widetilde{X}}^2})^{-1}),
    \label{xj_posterior_affine_generative}
    \\
    &p(\widetilde{x}_j|\widetilde{y}_j,\beta, \sigma^2) \stackrel{D}{=} \mathcal{N}(\widetilde{x}_j; \beta_1\widetilde{y}_j + \beta_0, \sigma^2)\label{xj_posterior_affine_discriminative}.
\end{align}
The Gibbs sampling procedure for semi-supervised learning of affine homoskedastic is summarized in algorithm \ref{alg:gibbs_affine_semi_supervised}. Again, the steps highlighted in red are indeed responsible for leveraging information from the unlabeled observations. 

\begin{algorithm}
\caption{Gibbs sampling from $p(x_0|y_0,\mathcal{D},$\textcolor{red}{$\mathcal{Y}$}$)$ using a generative (resp. discriminative) homoskedastic affine model}
\label{alg:gibbs_affine_semi_supervised}
\begin{algorithmic}
\REQUIRE{$ y_0,\mathcal{D}$, \textcolor{red}{$\mathcal{Y}$}, number of steps $T$}
\STATE{${\sigma^{2}}^{(0)} \sim \mathrm{I}\Gamma(\sigma^2;\lambda, \eta)$}
\STATE{$\beta^{(0)}\sim p(\beta|{\sigma^2}^{(0)},\mathcal{D})$ with \eqref{affine_coefficients_posterior_generative}}
\FOR{$t=1$ to  $T$}
\STATE{$x_0^{(t)} \sim p(x_0|y_0,\beta^{(t-1)}, {\sigma^2}^{(t-1)})$ with \eqref{x0_posterior_affine_generative} (resp. \eqref{affine_discriminative_model}}
\textcolor{red}{\FOR{all $\widetilde{y}_j \in \mathcal{Y}$}
\STATE{$\widetilde{x}_j^{(t)} \sim p(\widetilde{x}_j|\widetilde{y}_j,\beta^{(t-1)}, {\sigma^2}^{(t-1)})$ with \eqref{xj_posterior_affine_generative} (resp. \eqref{xj_posterior_affine_discriminative})}
\ENDFOR}
\STATE{set $\mathcal{D}_+^{(t)} = \mathcal{D} \cup (x_0^{(t)},y_0)$\textcolor{red}{$\cup\{(\widetilde{x}^{(t)}_j, \widetilde{y}_j)\}_{j=1}^{|\mathcal{Y}|}$}, ${\sigma^2}^{(t)} \sim p(\sigma^2|\beta^{(t-1)},\mathcal{D}_+^{(t)})$ with \eqref{noise_variance_posterior_generative} (resp. \eqref{noise_variance_posterior_discriminative})}
\STATE{$\beta^{(t)} \sim p(\beta|{\sigma^2}^{(t)},\mathcal{D}_+^{(t)})$ with \eqref{affine_coefficients_posterior_generative}(resp. \eqref{affine_coefficients_posterior_discriminative})}
\ENDFOR
\RETURN $x_0^{(T)}$
\end{algorithmic}
\end{algorithm}

\begin{figure}
    \centering
    \includegraphics[width = .8\textwidth]{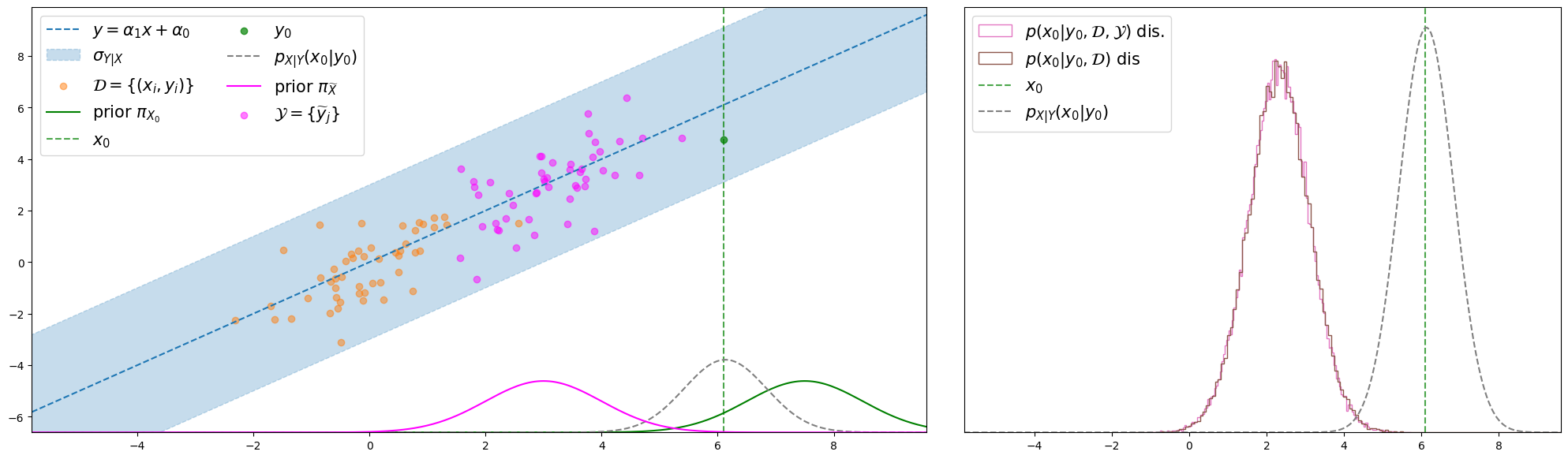}
    \caption{Samples from the supervised and semi-supervised Gibbs algorithm in affine discriminative modeling}
    \label{fig:discriminative_semi_supervised_result}
\end{figure}

We will now illustrate whether or not the modeling approach enables leveraging information in $\mathcal{Y}$ to contribute in reducing the epistemic uncertainty. We first provide empirical evidence that the discriminative approach is unable to leverage information in unlabeled observations to reduce the epistemic uncertainty. To that end, we consider a setting where $\mathcal{D}$ contains few points leading to high epistemic uncertainty to be reduced; and we apply the previous Gibbs sampling algorithm 
% broken reference \ref{alg:gibbs_affine_semi_supervised} 
SM3.1
in the discriminative case with and without unlabeled observations $\mathcal{Y}$ 
% \textcolor{blue}{removed here}
% (in the later case it reduces to algorithm \ref{alg:gibbs_affine_supervised})
and visually observe that the resulting samples seem to follow the same distribution. This empirical illustration is presented in figure \ref{fig:discriminative_semi_supervised_result}.
We further perform a Kolmogorov-Smirnof statistical test where the null hypothesis is $H_0:$ the samples obtained from the supervised and semi-supervised Gibbs sampling algorithm are from the same (unknown) distribution. This test works from two set of iid samples; but the samples from the Gibbs sampling algorithm are correlated samples so we first extract (almost) uncorrelated samples sub-sampling the Markov chain at every each integrated auto-correlation time steps. The Kolmogorov-Smirnof test yields a $p$-value of $0.566$ and we cannot reject the null hypothesis with low error probability; which indicates that the data i.e. the two sets of de-correlated samples is consistent with the null hypothesis i.e. that they originate from the same underlying probability distribution. 

Conversely, in the generative modeling approach, we go back to the setting presented in the previous figure \ref{fig:semi_supervised_affine_setting}, and we compare the supervised and semi-supervised ppds, from which we obtain samples via the corresponding Gibbs sampling algorithm
% \textcolor{blue}{removed here
% \ref{alg:gibbs_affine_supervised} and
% broken reference
\ref{alg:gibbs_affine_semi_supervised} 
with and without $\mathcal{Y}$. The results are presented in the next figure \ref{fig:semi_supervised_affine_result}. We compare the empirical distributions (built an histogram of the samples) and we visually observe that the semi-supervised ppd provides with a better of the true unknown posterior than the supervised ppd. To go further than a visual interpretation, we compare the calibration curves of the both ppds. The calibration curve allows to assess the quality of an approximation and is computed from the pdf of the target posterior, in our case the true unknown posterior and from samples from the approximating distribution, in our case the (supervised or semi-supervised) ppds. It is built by computing, for values $\alpha \in [0,1]$, the $\alpha$-highest density region of the target distribution using the pdf and computing the proportion of samples which land in that region. We can therefore conclude that the unlabeled observation were taken into account during the inference and indeed contributed to reducing the epistemic uncertainty.  
\begin{figure}[h!]
\begin{subfigure}{0.63\textwidth}
    \centering
    \includegraphics[width =\textwidth]{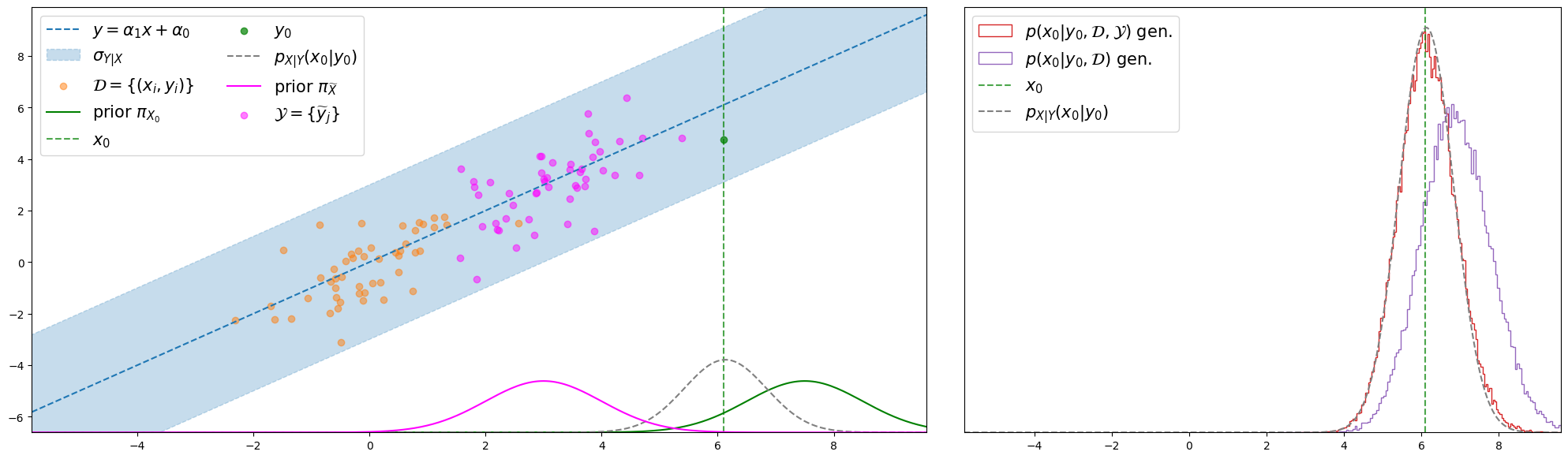}
    \caption{Unlabeled observations $\mathcal{Y}$ indeed contribute to reducing the generative epistemic uncertainty.}
    \label{fig:semi_supervised_affine_result}
\end{subfigure}
\hfill
\begin{subfigure}{0.32\textwidth}
    \centering
    \includegraphics[width =.6\textwidth]{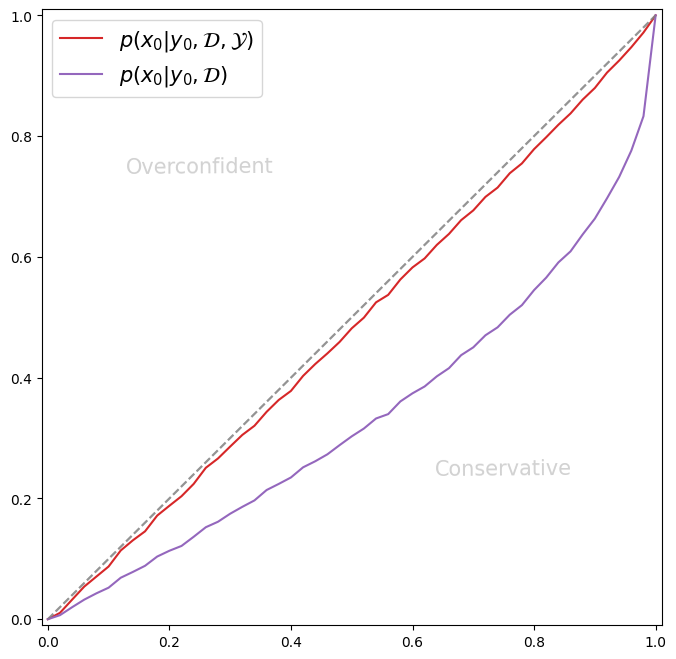}
    \caption{Calibration curves of supervised and semi-supervised ppds.}
    \label{fig:calibration}
\end{subfigure}
\caption{Semi-supervised generative affine modeling}
\end{figure}

\subsection{Parallel inference}\label{parallel_inference}

In this section we propose to re-discuss the problems of supervised and semi-supervised learning by considering them as solving two (or several) posterior inferences at the same time. To that end, we now denote $\mathcal{X} = \{\widetilde{x}_j|\widetilde{y}_j\sim \mathcal{P}_{Y|X}(Y|X = \widetilde{x}_j\}_{j=1}^{|\mathcal{Y}|}$ the set of unknown labels associated to the unlabeled observations. On the one hand, $\mathcal{X}$ are not necessarily label values of interest in the initial problem (that of inferring $x_0$ via $y_0$), it is nonetheless an unknown rv related to $\mathcal{Y}$ via the same unknown DGP and can be associated to an inference problem (again, possibly irrelevant in the context of the initial problem).
On the other hand, both $x_0$ and $\mathcal{X}$ can be values of interest. Indeed, in many learning instances, we dispose of a training dataset
(the set $\mathcal{D}$
in our paper)
to infer the probable models;
and given another set of observations
(the so-called testing dataset),
our aim is to predict 
for each of them the associated label
(which is different from one observation to another,
% and we wish to predict for several observations,
% each of them being associated 
% to a different label 
as opposed to section \ref{handling_mulitple_observations}, 
where one common label produced several observations). 
% This set of observations, together with the set of unknown labels, is often described as a testing dataset for evaluating model performance. 
In this case,
% as is the case with unlabeled observations,
all the observations in the testing dataset play an epistemic role in the generative case. More precisely, the observations of the testing dataset act as unlabeled observations resulting in an underlying semi-supervised learning setting. This is illustrated via quantitative simulations in the next section \ref{simulations}.

In this context, since we are trying to infer $x_0$ from $y_0$ and $\mathcal{X}$ from $\mathcal{Y}$ via probable models $\theta$ using the same observations $\mathcal{D}$, the two inference should not be treated as independent problems in both modelling approaches. However, a main difference between the two modeling approaches can be understood when considering two (or several) inference problems at once.
% \begin{eqnarray}
%     \int p(x_0, \mathcal{X}|y_0,\mathcal{Y}, \mathcal{D}) \mathrm{d}\mathcal{X} \stackrel{G}{=} p(x_0|y_0,\mathcal{Y},\mathcal{D}) \stackrel{G}{\neq} p(x_0|y_0,\mathcal{D}),
%     \\
%     \int p(x_0, \mathcal{X}|y_0,\mathcal{Y}, \mathcal{D}) \mathrm{d}{x_0} \stackrel{G}{=} p(\mathcal{X}|\mathcal{Y},y_0,\mathcal{D}) \stackrel{G}{\neq} p(\mathcal{X}|\mathcal{Y},\mathcal{D}),
% \end{eqnarray}

 In the discriminative setting, we have that $p(x_0, \mathcal{X}|y_0,\mathcal{Y},\mathcal{D}) \stackrel{D}{=}p(x_0|y_0,\mathcal{D})p( \mathcal{X}|\mathcal{Y},\mathcal{D})$
so the inference of $x_0$ (resp $\mathcal{X}$) does not depend on $\mathcal{Y}$ (resp $y_0$). As such, each inference problem can be solved via sampling from its corresponding ppd. Conversely, in the generative setting, all the observations $y_0,\mathcal{Y}$ act as unlabelled observations in both inferences and the two problems should not be treated independently as all the unlabeled observations contribute to reducing the epistemic uncertainty. As a result, $p(x_0, \mathcal{X}|y_0,\mathcal{Y},\mathcal{D}) \stackrel{G}{\neq}p(x_0|y_0,\mathcal{D})p( \mathcal{X}|\mathcal{Y},\mathcal{D})$ and both inference can not be solved by sampling from each corresponding ppd, and this modelling approach instead calls for a sampling from the joint distribution. 
Finally, note that the previous semi-supervised algorithm, which we proposed as a way to obtain samples from the posterior $p(x_0|y_0,\mathcal{D},\mathcal{Y})$ was constructed by applying the Gibbs sequential sampling mechanism to the joint distribution $p(x_0, \mathcal{X}, \theta|y_0,\mathcal{Y}, \mathcal{D})$ and as such, produced desired samples from $p(x_0|y_0,\mathcal{D},\mathcal{Y})$ but also, as a byproduct, samples from $p(\mathcal{X}|\mathcal{Y}, \mathcal{D},y_0)$, effectively solving both inference problems at once. Again of course, since the Gibbs is only a tool for solving the inference problem, the underlying structure of dependency between rv is preserved. So in the discriminative case these distributions respectively reduce to $p(x_0|y_0,\mathcal{D})$ and $p(\mathcal{X}|\mathcal{Y}, \mathcal{D})$.

% In many learning instances, we dispose of a training dataset
% (the set $\mathcal{D}$
% in our paper)
% to infer the probable models;
% and given another set of observations
% (the so-called testing dataset),
% our aim is to predict 
% for each of them the associated label
% (which is different from one observation to another,
% % and we wish to predict for several observations,
% % each of them being associated 
% % to a different label 
% as opposed to section \ref{handling_mulitple_observations}, 
% where one common label produced several observations). 
% % This set of observations, together with the set of unknown labels, is often described as a testing dataset for evaluating model performance. 
% In this case,
% % as is the case with unlabeled observations,
% all the observations in the testing dataset play an epistemic role in the generative case. More precisely, the observations of the testing dataset act as unlabeled observations resulting in an underlying semi-supervised learning setting. This is illustrated via quantitative simulations in the next section \ref{simulations}.

\section{Simulations}\label{simulations}
Throughout the paper, we leveraged the example of affine homoskedastic modeling in the case of univariate regression to illustrate the arguments of this paper. As we mentionned before, this illustrating example can be relevant to some readers as modeling affine dependencies is a most frequent problem in many scientific fields. 
We also used 
%Moreover, we primarily used 
this example as it enables, with appropriate choice of priors, to run a straightforward Gibbs sampler for both the generative and discriminative modeling approaches.
% , which enabled us to illustrate the arguments of this paper via an exact sampling
However, considering models which enable such convenient sampling procedures with closed-form $p(\theta|\mathcal{D})$ (or all its conditionals in a Gibbs scheme) indeed heavily restricts the choice of model $\mathcal{P}_\theta$. In this section we now consider conditional models $\mathcal{P}_\theta$ which are defined using NN functions with tractable pdf, but for which we are not able to elicit a prior $\Pi_\theta$ over parameters $\theta$ such that the posterior distribution $p(\theta|\mathcal{D})$ admits a closed form expression, and we resort to approximate sampling from that posterior using Stochastic Gradient Langevin Dynamics \cite{teh2016consistency} \cite{durmus2017nonasymptotic}. 

In this section, we tackle the problem of classification in which $X$ is hence a Categorical rv. We evaluate generative and discriminative models both defined via NN functions (we describe the specific structure hereafter) and with similar number of parameters. We proceed to assess the classification accuracy of a generative versus discriminative model for $\theta \sim p(\theta|\mathcal{D},y_0)$ via Gibbs sampling. We consider three different scenarios which we now describe. 

Following the idea described in section \ref{parallel_inference}, we now consider 
two distincts sets:
% $\mathcal{D} \stackrel{\Delta}{=} \{(x_i, y_i)|x_i \sim \mathcal{P}^\mathcal{D}_{X}, y_i\sim \mathcal{P}_{Y|X}(Y|X = x_i)\}_{i=1}^{|\mathcal{D}|}$
the training dataset $\mathcal{D} =
\{ (x_{i},y_{i}) \}_{i=1}^N$,
and a testing dataset
$\{ (x_{0,j},y_{0,j}) \}_{j=1}^M$.
% in which 
% %comprised of a set of unknown labels of interest
% $x_0 = \{x_{0,j}\}$ is the set of labels and 
% $y_0 = \{y_{0,j}\}$ 
% the set of associated observations. 
All the labels $x_{0,j}$ 
are distributed according to $\Pi_{X_0}$, and in the following we address three scenarios,
which differ on 
the size of the testing set,
and the possible discrepancy between 
$\Pi_{X_0}$ 
and
$\mathcal{P}_{X}^{\mathcal{D}}$.
% well as the distribution of labels $\Pi_{X_0}$ compared to $\mathcal{P}_{X}^{\mathcal{D}}$ will vary.

% we now consider two distinct sets. 
% $\mathcal{D}$ describes the set of labeled observations, often otherwise referred to as the training set. In this section, and by contrast with section \ref{handling_mulitple_observations} where a single label produced several observations, 
% we now consider $x_0 = \{x_{0,j}\}$ to be a set of labels of interest,
% in which 
% each $x_{0,j}$ is drawn from , 
% and is associated with one observation $y_{0,j}$.
% Finally let 
% $y_0 = \{y_{0,j}\}$. 

\textbf{Scenario 1: 
Identical priors and sizes.}
% $N=M$ and $\mathcal{P}_X^{\mathcal{D}} = \Pi_{X_0}$}. 
We first consider the scenario where the label distribution from the training dataset 
coincides with 
%is the same as 
the prior. 
So $\mathcal{P}_X^{\mathcal{D}} = \Pi_{X_0}$,
and as such, couples $(x_i, y_i)$ (which belong to training dataset $\mathcal{D}$) and $(x_{0,j}, y_{0,j})$ (which belong to the testing dataset) have the same distributions. This corresponds to the most frequent situation in practice and we use this setting as a baseline. We consider the dataset $\mathcal{D}$ 
and 
the set $\{y_{0,j}\}$ 
of unlabeled observations to be of the same size,
i.e. $N=M$.

\begin{wrapfigure}{R}{4.7cm}
\centering
\includegraphics[width=.3\textwidth]{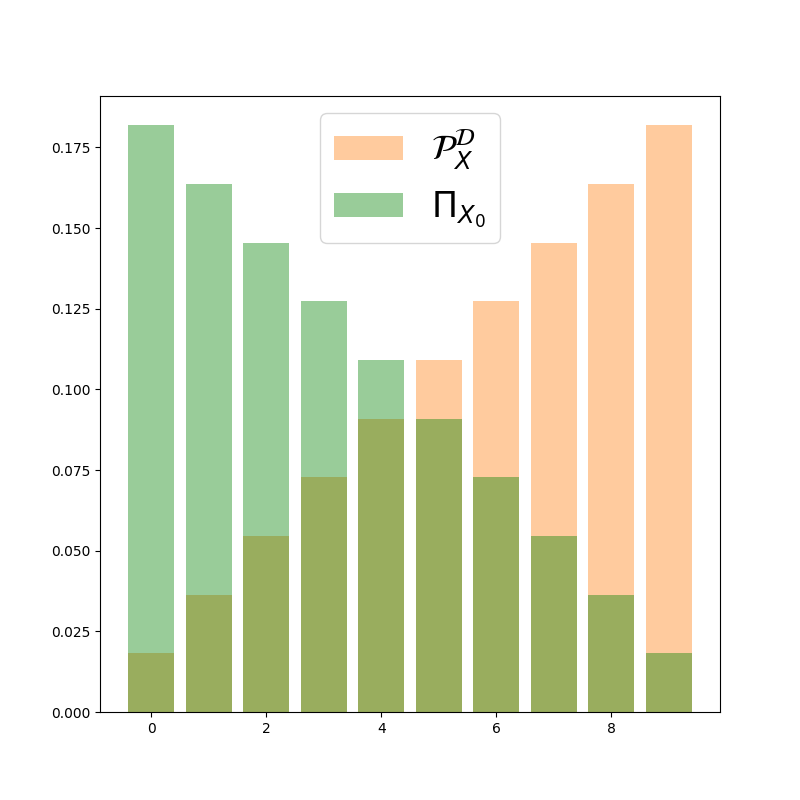}
\caption{Different prior distributions in scenario 2.}\label{fig:priors_scenario_2}
\end{wrapfigure}

\textbf{Scenario 2: Imbalanced dataset (different priors, same sizes)}. 
We then consider the scenario where 
$\mathcal{P}_X^{\mathcal{D}}
\neq \Pi_{X_0}$, 
so couples 
$(x_i, y_i)$
and
$(x_{0,j}, y_{0,j})$ 
do not have the same distributions
(and indeed are quite different - 
see figure \ref{fig:priors_scenario_2}).
% To achieve this situation in practice, we proceed to random subsampling such that the $x_i$ values of the dataset are drawn according to $\mathcal{P}_X^{\mathcal{D}}$. 
%We display the frequencies of labels in $\mathcal{D}$ and that of the unknown labels in the next figure. 
%We still consider the dataset $\mathcal{D}$  and 
%the set $\{y_{0,j}\}$ 
%of unlabeled observations to be of the same size,
%i.e. $N=M$.
We still set $N=M$.

% We first consider the scenario where the label distribution from the training dataset 
% coincides with 
% %is the same as 
% the prior. 
% So $\mathcal{P}_X^{\mathcal{D}} = \Pi_{X_0}$,
% and as such, couples $(x_i, y_i)$ (which belong to training dataset $\mathcal{D}$) and $(x_{0,j}, y_{0,j})$ (which belong to the testing dataset) have the same distributions. This corresponds to the most frequent situation in practice and we use this setting as a baseline. We consider the dataset $\mathcal{D}$ 
% and 
% the set $\{y_{0,j}\}$ 
% of unlabeled observations to be of the same size,
% i.e. $N=M$.

\textbf{Scenario 3: Few labeled samples
(same priors, different sizes)}. 
We finally consider the scenario where $\mathcal{P}_X^{\mathcal{D}}
=
\Pi_{X_0}$,
but $N\ll M$,
so we dispose of a few labeled observations in $\mathcal{D}$, and of a large amount of observed values $y_{0,j}$ for which we want to infer the corresponding label $x_{0,j}$.
In this setting, the low number of observations in $\mathcal{D}$ will hinder the prediction accuracy of both models, but as we have explained before, the large amount of unlabeled observations act as an unlabeled dataset in the generative case.

We consider both the classification datasets of MNIST and of FashionMNIST, for which we reduce the dimension of observations via a Principal Component Analysis \cite{pearson1901liii} in order to keep $95\%$ of explained variation. We compare a discriminative model which is a fully connected NN with 4 hidden-layers of 256 units to a generative model which is built using a combination of invertible conditional Normalizing Flows layers \cite{dinh2016density} and stochastic ones \cite{argouarch2022discretely}. We sample from the joint distribution via Gibbs sampling with $T = 10$ steps.
% , and at each step we sample from the posterior over model parameters using Stochastic Gradient Langevin Dynamics.
% We provide all reproducible code and experiments in the Github repository at \href{https://github.com/ElouanARGOUARCH/Generative_Discriminative_Uncertainty_Quantification}{github.com/ElouanARGOUARCH/Generative\_Discriminative\_Uncertainty\_Quantification}.
The results are provided in the next table \ref{table:classif_accuracy} and for each dataset and scenario, we consider 10 independent runs and we display
the average classification accuracy as well as the standard deviation.
\begin{table}[!ht]
\centering
\begin{tabular}{ |c||c|c||c|c| } 
\hline
Dataset&\multicolumn{2}{c||}{MNIST}&\multicolumn{2}{c|}{FashionMNIST}\\
\hline
Model. & Disc. & Gen. & Disc. & Gen. \\
\hline
Scenario 1  & $0.9628 \pm 0.0014$ & $\mathbf{0.9749\pm 0.0145}$ &$\mathbf{0.8767\pm 0.0009}$ &$0.8652\pm0.0094$
\\ 
Scenario 2 & $0.9520 \pm  0.0007$ & $\mathbf{0.9774\pm 0.0213}$ &$0.7000\pm0.0057$ & $\mathbf{0.8482\pm 0.0172}$
\\ 
Scenario 3 &$0.9349 \pm  0.0018$ & $\mathbf{0.9690\pm 0.0237}$ &$0.7618\pm0.0008$ & $\mathbf{0.8365\pm0.0187}$\\ 
\hline
\end{tabular}
\caption{Results of classification accuracy in percentages of Discriminative and Generative modeling approaches}\label{table:classif_accuracy}
\end{table}

We now analyze the results of this experiment. Comparing the results for the first scenario tells us that the generative modeling can indeed be on par in terms of classification accuracy when compared to its discriminative counterpart. This scenario can be used as a baseline experiment for the two following scenarios. In the second scenario, the distribution of labels in the dataset, $\mathcal{P}_X^\mathcal{D}$, is different from $\Pi_{X_0}$, leading to a situation of imbalanced dataset. 
In this situation, we notice that the discriminative model indeed suffers in term of accuracy as it favors the dominant classes of the dataset. As we have explained, the generative approach does not suffer from such dataset imbalance, or at least not as much, which is confirmed in this experiment. This is in accordance with the discussion of section \ref{explicit_implicit_prior}. Finally, in the third scenario, the lower number of labeled observation in $\mathcal{D}$ hinders, as expected, the classification accuracy of the discriminative model as compared to its generative counterpart,
as the latter indeed leverages the unlabeled observations in the inference of probable models, which indeed contributes to reduce the modeling epistemic uncertainty,
as discussed in section \ref{semi supervised with gen vs dis}. 

\section*{Conclusion}

Throughout this paper, we discussed the epistemic uncertainty quantification in generative and discriminative models,
and draw several conclusions. On the one hand,
discriminative models are an easy-to-use tool since they can be parameterized easily and directly approximate the posterior.
Moreover, if they can be sampled from easily, 
then one can use a straightforward two step procedure for sampling
from the ppd,
which indeed enables to quantify the epistemic uncertainty.
However, by nature
discriminative models 
do not take into account 
the information contained in the prior distribution, which is replaced by an implicit prior inferred on the dataset. 
As a result, they suffer from imbalanced datasets. 
Finally they cannot be conveniently used in the context of inferring from multiple observations, and they cannot leverage information from unlabeled data.

On the other hand, generative models are perhaps less convenient to use as they usually require a more sophisticated structure and require an additional inference step, in addition to the prior distribution, to sample from the corresponding posterior. 
Yet by construction they do enable to leverage information from all available sources,
making them an appealing tool, in particular in a semi-supervised context.
In practice, the two-step procedure for sampling from the ppd 
is no longer available; 
but our general purpose Gibbs sampling based algorithm indeed enables to
sample from the ppd (and thus perform epistemic uncertainty quantification) while taking into account both the labeled and unlabeled observations.

\bibliographystyle{plain.bst} 
\bibliography{bibliography.bib} 

\end{document}